# Sherlock: Scalable Fact Learning in Images

Mohamed Elhoseiny[1,2], Scott Cohen[1], Walter Chang[1], Brian Price[1], Ahmed Elgammal[2]

[1]Adobe Research          [2]Department of Computer Science, Rutgers University

**Abstract.** We study scalable and uniform understanding of facts in images. Existing visual recognition systems are typically modeled differently for each fact type such as objects, actions, and interactions. We propose a setting where all these facts can be modeled simultaneously with a capacity to understand unbounded number of facts in a structured way. The training data comes as structured facts in images, including (1) objects (e.g., <boy>), (2) attributes (e.g., <boy, tall>), (3) actions (e.g., <boy, playing>), and (4) interactions (e.g., <boy, riding, a horse >). Each fact has a semantic language view (e.g., < boy, playing>) and a visual view (an image with this fact). We show that learning visual facts in a structured way enables not only a uniform but also generalizable visual understanding. We propose and investigate recent and strong approaches from the multiview learning literature and also introduce two learning representation models as potential baselines. We applied the investigated methods on several datasets that we augmented with structured facts and a large scale dataset of more than 202,000 facts and 814,000 images. Our experiments show the advantage of relating facts by the structure by the proposed models compared to the designed baselines on bidirectional fact retrieval.

## 1 Introduction

Despite recent significant advances in recognition, image captioning, and visual question answering (VQA), there is still a large gap between humans and machines in the deep image understanding of objects, their attributes, actions, and interactions with one another. The human visual system is able to efficiently gain visual knowledge by learning different types of facts in a never ending way from many or few examples, aided by the ability to generalize from other known facts with related structure. We believe that the most effective and fastest way to close this gap are with methods that possess that following key characteristics:

- **Uniformity**: The method should be able to handle objects ("dog"), attributes ("brown dog"), actions ("dog running") and interactions between objects ("dog chasing cat").
- **Generalization**: The method should be able to generalize to facts that have zero or few examples during training.
- **Scalability**: The method should handle an unbounded number of facts.
- **Bi-directionality**: The method should be able to retrieve a language description for an image, and images that show a given language description of a fact.
- **Structure**: The method should provide a structured understanding of facts, for example that "dog" is the subject and has an attribute of "smiling".

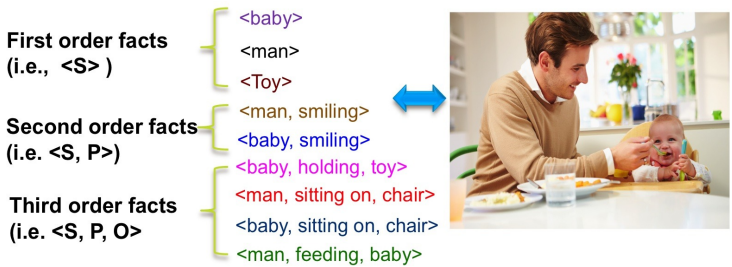

Fig. 1: Visual Facts in Images

Existing visual understanding systems may be categorized into two trends: (1) fact-level systems and (2) high-level systems. Fact level systems include object recognition, action recognition, attribute recognition, and interaction recognition (e.g., [36], [43], [6], [44], [14], [3]). These systems are usually evaluated separately for each fact type (e.g., objects, actions, interactions, attributes, etc.) and are therefore not uniform. Typically, these systems have a fixed dictionary of facts, assuming that facts are seen during training by at least tens of examples, and treat facts independently. Such methods cannot generalize to learn facts outside of the dictionary and will not scale to an unbounded number of facts, since model size scales with the number of facts. Furthermore, these recognition systems are typically uni-directional, only able to return the conditional probability of a fact given an image. The zero/few-shot learning setting (e.g., [33,23]), where only a few or even zero examples per fact are available, is typically studied apart from the traditional recognition setting. We are not aware of a unified recognition/few shot learning system that learns unbounded set of facts.

In the second trend, several researchers study tasks like image captioning [19,39,40,25], image-caption similarity [19,21], and visual question answering [2,24,32] with very promising results. These systems are typically learning high-level tasks but their evaluation does not answer whether these systems relate captions or questions to images by fact-level understanding. Captioning models output sentences and thus can mention different types of facts and, in principle, any fact. However, Devlin et al. [8,9] reported that 60-70% of the generated captions by LSTM-based captioning methods actually exist in the training data and show that nearest neighbor methods have very competitive performance in captioning. These results call into question both the core understanding and the generalization capabilities of the state-of-the-art caption-level systems.

The limitations of prior settings motivated us to study a fact-level understanding setting, which is more related to the first trend but unified to any fact type and able to learn an unbounded number of facts. This setting allows measuring the gained visual knowledge represented by the facts learnt by any proposed system to solve this task. Our goal is a method that achieves a more sophisticated understanding of the objects, actions, attributes, and interactions between objects, and possesses the desireable properties of scalability, generalization, uniformity, bi-directionality, and structure.

Our approach is to learn a common embedding space in which the language and visual views of a fact are mapped to the same location. The key to our solution achieving the desireable characteristics is to make the basic unit of understanding a structured fact as shown in Fig. 1 and to have a structured embedding space in which different dimensions record information about the subject S, predicate P, and object O of a fact.

Using an embedding space approach allows our method to scale as we can submit any (<S,P,O>, image), (<S,P>, image), or (<S>, image) facts to train our embedding network. At test time, it allows for bi-directional retrieval, as we can search for language facts that embed near a given image fact and vice-versa. Retaining the structure of a fact in the embedding space gives our method the chance to generalize to understand an S/SP/SPO from training data on its S, P, and O components, since this information is kept separate. To obtain uniformity, we introduce wildcards "*" into our structured fact representation, e.g. <man,smiling,*> or <dog,*,*> and use a wildcard training loss which ignores the unspecified components of embedded second and first order visual and language facts. Carefully designed experiments show that our uniform method achieves state-of-the-art performance in fact-level bidirectional view retrieval over existing image-sentence correlation methods, other view embedding methods, and a version of our method without structure, while also scaling and generalizing better.

**Contributions:** (1) We propose a new problem setting to study fact-level visual understanding of unbounded number of facts while considering the aforementioned characteristics. (2) We design and investigate several baselines from the multiview learning literature and apply them on this task. (3) We propose two learning representation models that relate different fact types using the structure exemplified in Fig 1. (4) Both the designed baselines and the proposed models embed language views and visual views (images) of facts in a joint space that allows uniform representation of different fact types. We show the value of relating facts by structure in the proposed models compared to the designed baselines on several datasets on bi-directional fact retrieval.

## 2   Related Work

In order to make the contrast against the related work clear, we start by stating the scale of facts we are modeling in this work. Let's assume that |S|, |P|, and |O| denotes the number of unique subjects, unique predicates, and unique objects, respectively; see Fig 1. The scale of unique second and third order facts is bounded by $|S| \times |P|$ and $|S| \times |P| \times |O|$ possibilities respectively, which can easily reach millions of facts. The data we collected in this work has thus far reached 202,000 unique facts (814,000 images). We cover five lines of related research (first three are from fact-level recognition literature).

**(A) Modeling Visual facts in Discrete Space:** Recognition of objects or activities has been typically modeled as a mapping function $g : \mathcal{V} \to \mathcal{Y}$, where $\mathcal{Y}$ is discrete set of classes. The function $g$ has recently been learned using deep learning (e.g., [36,38]). Different systems are built to recognize each fact type in images by modeling a different $g : \mathcal{V} \to \mathcal{Y}$ , where $\mathcal{Y}$ is constrained to objects, (e.g., [36]), attributes (e.g. [43]), attributed objects (<car, red>) [6], scenes (e.g., [44]), human actions (e.g., [14]), and interactions [3]. There are several limitations for modeling recognition as $g : \mathcal{V} \to \mathcal{Y}$ with $|\mathcal{Y}| \to \infty$. **(1) Scalability:** Adding a new fact leads to changing the architecture, meaning adding thousands of parameters and re-training the model (e.g., for adding a new output node). For example, if VGGNet [36] is used on the scale of 202,000 facts, the number of parameters in the softmax layer alone is close to 1 billion. **(2) Uniformity:** Modeling each group of facts by a different $g$ requires maintaining different systems, retrain several models as new facts are added, and also doesn't allow learning the correlation among different fact types. However, we aim to uniformly model visual perception. **(3) Generalization:** While most of the existing benchmarks for this setting have at least tens of examples per fact (e.g., imageNet [7]), a more realistic assumption is that there might not be enough examples to learn the new class (the long-tail problem). Several works have been proposed to deal this problem in object recognition settings [45,35]. However, they suffer from the aforementioned scalability problems as facts increase. **(4) Bi-directionality:** These models are uni-directional from $\mathcal{V}$ to $\mathcal{Y}$. Fig 2 shows representatives settings of these methods. The three axes are Scalability, Uniformity, and Generalization. These methods typically study seen classes and hence do not generalize to unseen classes.

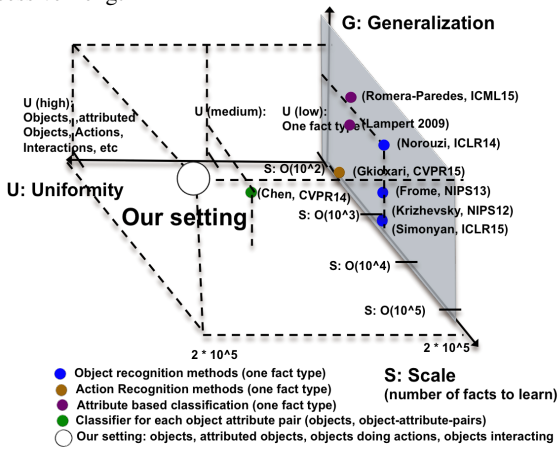

Fig. 2: Our setting in contrast to the studied fact recognition settings in the literature. Scalability means the number of facts studied in these works. Uniformity means if the setting is applied for multiple fact types. Generalization means the performance of this methods on facts of zero/few images.

**(B) Modeling zero/few shot fact learning by semantic representation of classes (e.g., attributes):** One of the most successful ideas for learning from few examples per class is by using semantic output codes like attributes as an intermediate layer between features and classes. Formally, $g$ is a composition of two function $g = h(a(\cdot))$, where $a : \mathcal{V} \to \mathcal{A}$, and $h : \mathcal{A} \to \mathcal{Y}$ [30]. The main idea is to collect data that is sufficient to learn an intermediate attribute layer, where classes are then represented by these attributes to facilitate zero-shot/few-shot learning. However, Chen *et al*. [6] realized that attribute appearance is dependent on the class, as opposed to these earlier models [30,23,12]. Although [6]'s assumption is more realistic, they propose learning different classifiers for each category-attribute pair, which suffers from the same scalability and learning problems pointed out in (A) and is restricted to certain groups of facts (not uniform).

More recent attribute-based zero-shot learning methods embed both images and attributes into a shared space (e.g., Attribute Embedding [1], ESZSL [33]). These methods were mainly studied in the case of zero-shot learning and have shown strong performance. In contrast, we aim at studying the setting where one system that can learn from both facts with many training images and facts with few/no training images. Fig 2 shows the contrast between our setting (white circle) and this setting. Although these methods were mainly studied using attributes as a semantic representation and at a much smaller scale of facts, we apply the state of the art ESZSL [33] in order to study the capacity of these models at a much larger scale.

**(C) Object Recognition in continuous space using Vision and Language:** Recent works in language and vision involve using unannotated text to improve object recognition and to facilitate zero-shot learning. The following group of approaches model object recognition as a function $g(v) = \arg\max_y s(v \in \mathcal{V}, y \in \mathcal{Y})$, where $s(\cdot, \cdot)$ is a similarity function between image $v$ and class $y$ represented by text. In [13], [29] and [37], word embedding language models (e.g., [26]) were adopted to represent class names as vectors. In their setting, the imageNet dataset has 1000 object facts with thousands of examples per class. Our setting has two orders of magnitude more facts with a long-tail distribution. Conversely, other works model the mapping of unstructured text descriptions for classes into a visual classifier [10,5]. We are extending the visual recognition task to unbounded scale of facts, not only object recognition but also attributes, actions, and interactions in one model; see Fig 2 for contrast to our setting.

**(D) Image-Caption Similarity Methods:** As we illustrated earlier, our goal is fact-level understanding. However, image-caption similarity methods such as [19,21] are relevant as multi-view learning methods. Although it is a different setting, we found two interesting aspects of these methods to study in our setting. First, how image-caption similarity system trained on image-caption level performs on fact-level understanding. Second, these systems could be retrained in our setting by providing them with fact-level annotation, where every example is a phrase representing the fact and an image (e.g., "person riding horse" and an image with this fact).

**(E) MV-CCA :** MV-CCA is a recent multiview, scalable, and robust version of the famous CCA embedding method [15]. We apply MV-CCA as a baseline in our setting.

## 3  Problem Definition: Representation and Visual Modifiers

We deal with three groups of facts; see Fig. 1. First Order Facts <S,*,*> are object and scene categories (e.g., <baby,*,*>, <girl,*,*>, <beach,*,*>). Second Order Facts <S,P,*> are objects performing actions or attributed objects (e.g., <baby, smiling,*>, <baby, Asian,*>). Third Order Facts <S,P,O> are interactions and positional information (e.g. <baby, sitting_on, high_chair>, <person, riding, horse>). By allowing wild-cards in this structured representation (<baby,*,*>and <baby, smiling,*>), we can not only allow uniform representation of different fact types but also relate them by structure. We propose to model these facts by embedding them into a structured fact space that has three continuous hyper-dimensions $\phi_S$, $\phi_P$, and $\phi_O$

$\phi_S \in \mathbb{R}^{d_S}$: The space of object categories or scenes S.
$\phi_P \in \mathbb{R}^{d_P}$: The space of actions, interactions, attributes, and positional relations.
$\phi_O \in \mathbb{R}^{d_O}$: The space of interacting objects, scenes that interact with S for SPO facts.

where $d_S$, $d_P$, and $d_O$ are the dimensionalities corresponding to $\phi_S$, $\phi_P$, and $\phi_O$, respectively. As shown in Fig. 3, first order facts like <woman,*,*>, <man,*,*>, <person,*,*> live in a hyper-plane in the $\phi_P \times \phi_O$ space. Second order facts (e.g., <man, walking,*>, <girl, walking,*>) live as a hyper-line that is parallel to $\phi_O$ axis. Finally, a third order fact like <man, walking, dog> is a point in the $\phi_S \times \phi_P \times \phi_O$ visual perception space. Inspired from the concept of language modifiers, the $\phi_S$, $\phi_P$, and $\phi_O$ could be viewed as what we call "visual modifiers". For example, the second order fact <baby, smiling,* > is a $\phi_P$ visual modifier for <baby,*,*>, and the third order fact <person, playing, flute> is the fact <person, *, *> visually modified on both $\phi_P$ and $\phi_O$ axes. By embedding all language and images into this common space, our algorithm can scale efficiently. Further, this space can be used to retrieve a language view of an image as well as a visual view of a language description, making the model bi-directional.

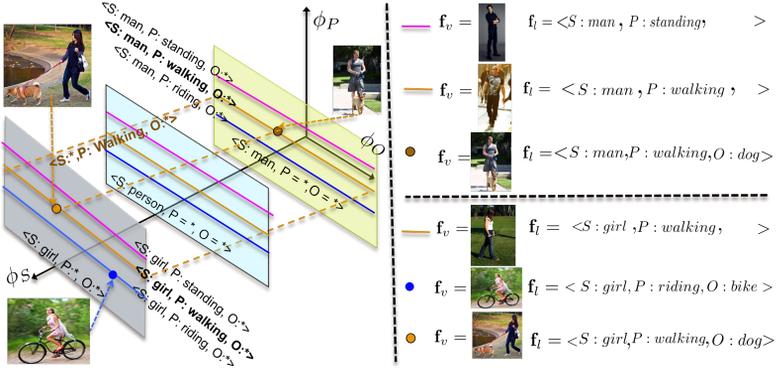

Fig. 3: Unified Fact Representation and Visual Modifiers Notion

We argue that modeling visual recognition based on this notion gives it a generalization capability. For example is if the model learned the facts <boy>, <girl>, <boy, petting, dog>, <girl, riding, horse>, we would aim at recognizing an unseen fact <boy, petting, horse>. We show these capabilities quantitatively later in our experiments. We model this setting as a problem with two views, one in the visual domain $\mathcal{V}$ and one in the language domain $\mathcal{L}$. Let $\mathbf{f}$ be a structured fact, $\mathbf{f}_v \in \mathcal{V}$ denoting the visual view of $\mathbf{f}$ and $\mathbf{f}_l \in \mathcal{L}$ denoting the language view of $\mathbf{f}$. For instance, an annotated fact with language view $\mathbf{f}_l = $<S:girl, P:riding, O:bike> would have a corresponding visual view $\mathbf{f}_v$ as an image where this fact occurs; see Fig. 4.

Our goal is to learn a representation that covers all the three orders of facts. We denote the embedding functions from a visual view to $\phi_S$, $\phi_P$, and $\phi_O$ as $\phi_S^{\mathcal{V}}(\cdot)$, $\phi_P^{\mathcal{V}}(\cdot)$, and $\phi_O^{\mathcal{V}}(\cdot)$, and the structured visual embeddings of a fact $\mathbf{f}_v$ by $\mathbf{v}_S = \phi_S^{\mathcal{V}}(\mathbf{f}_v)$, $\mathbf{v}_P = \phi_P^{\mathcal{V}}(\mathbf{f}_v)$, and $\mathbf{v}_O = \phi_O^{\mathcal{V}}(\mathbf{f}_v)$, respectively. Similarly, we denote the embedding functions from a language view to $\phi_S$, $\phi_P$, and $\phi_O$ as $\phi_S^{\mathcal{L}}(\cdot)$, $\phi_P^{\mathcal{L}}(\cdot)$, and $\phi_O^{\mathcal{L}}(\cdot)$, and the structured language embeddings of a fact $\mathbf{f}_l$ as $\mathbf{l}_S = \phi_S^{\mathcal{L}}(\mathbf{f}_l)$, $\mathbf{l}_P = \phi_P^{\mathcal{L}}(\mathbf{f}_l)$, and $\mathbf{l}_O = \phi_O^{\mathcal{L}}(\mathbf{f}_l)$.

We denote the concatenation of the visual view hyper-dimensions' embedding as $\mathbf{v}$, and the language view hyper-dimensions' embedding as $\mathbf{l}$; see Eq. 1 Third-order facts <S,P,O> can be directly embedded in the structured fact space by Eq. 1 with $\mathbf{v} \in \mathbb{R}^{d_S} \times \mathbb{R}^{d_P} \times \mathbb{R}^{d_O}$ for the image view and $\mathbf{l} \in \mathbb{R}^{d_S} \times \mathbb{R}^{d_P} \times \mathbb{R}^{d_O}$ for the language view. Based on our "fact modifier" observation, we propose to represent both second and first-order facts as wild cards "$*$", as illustrated in Eq. 2, 3; see Fig 4, 3.

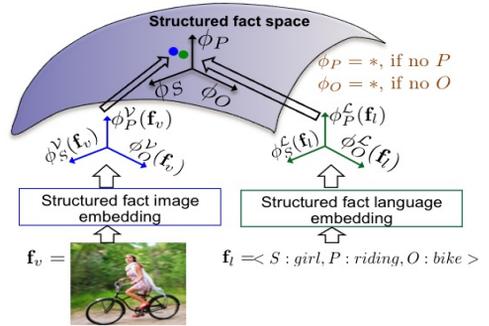

Fig. 4: Structured Embedding

**Third-Order Facts <S,P,O>:** $\mathbf{v} = [\mathbf{v}_S, \mathbf{v}_P, \mathbf{v}_O]$     $\mathbf{l} = [\mathbf{l}_S, \mathbf{l}_P, \mathbf{l}_O]$     (1)

**Second-Order Facts <S, P,*>:** $\mathbf{v} = [\mathbf{v}_S, \mathbf{v}_P, \mathbf{v}_O = *]$     $\mathbf{l} = [\mathbf{l}_S, \mathbf{l}_P, \mathbf{l}_O = *]$     (2)

**First-Order Facts <S,*,*>:** $\mathbf{v} = [\mathbf{v}_S, \mathbf{v}_P = *, \mathbf{v}_O = *]$     $\mathbf{l} = [\mathbf{l}_S, \mathbf{l}_P = *, \mathbf{l}_O = *]$     (3)

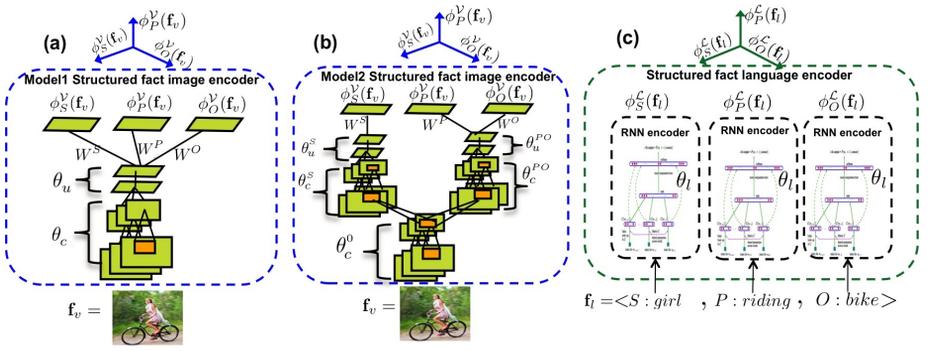

Fig. 5: Sherlock Models. See Fig. 4 for the full picture.

Setting $\phi_P$ and $\phi_O$ to $*$ for first-order facts means that the $P$ and $O$ modifiers are not of interest for first-order facts, which is intuitive. Similarly, setting $\phi_O$ to $*$ for second-order facts indicates that the $O$ modifier is not of interest for single-frame actions and attributed objects. If an image contains lower order fact such as <man>, then higher order facts such as <man, tall> or <man, walking, dog> may also be present. Hence, the wild cards (i.e. $*$) of the first- and second-order facts are not penalized during training.

## 4  Models

We propose a two-view structured fact embedding model with five properties mentioned in Sec 1. Satisfying the first four properties can be achieved by using a generative model $p(\mathbf{f}_v, \mathbf{f}_l)$ that connects the visual and the language views of $\mathbf{f}$, where more importantly $\mathbf{f}_v$ and $\mathbf{f}_l$ inhabit a continuous space. We model $p(\mathbf{f}_v, \mathbf{f}_l) \propto s(\mathbf{v}, \mathbf{l})$, where $s(\cdot, \cdot)$ is a similarity function defined over the structured fact space. We satisfy the fifth property by building our models over the aforementioned structured wild card representation. Our objective is that two views of the same fact should be embedded so that they are close to each other; see Fig 4. The question now is how to model and train $\phi^{\mathcal{V}}(\cdot)$ visual functions $(\phi_S^{\mathcal{V}}(\cdot), \phi_P^{\mathcal{V}}(\cdot), \phi_O^{\mathcal{V}}(\cdot))$ and $\phi^{\mathcal{L}}(\cdot)$ language functions $(\phi_S^{\mathcal{L}}(\cdot), \phi_P^{\mathcal{L}}(\cdot), \phi_O^{\mathcal{L}}(\cdot))$ . We model $\phi^{\mathcal{V}}(\cdot)$ as a CNN encoder (e.g., [22,36]), and $\phi^{\mathcal{L}}(\cdot)$ as RNN encoder (e.g., [26,31]) due to their recent success as encoders for images and words, respectively. We propose two models for learning facts, denoted by Model 1 and Model 2. Both models share the same structured fact language embedding/encoder but differ in the structured fact image encoder.

We start by defining an activation operator $\psi(\theta, a)$, where $a$ is an input, and $\theta$ is a series of one or more neural network layers (may include different layer types, e.g., convolution, pooling, then another convolution and pooling). The operator $\psi(\theta, a)$ applies $\theta$ parameters layer by layer to compute the final activation of $a$ using $\theta$ subnetwork.

**Model 1 (structured fact CNN image encoder):** In Model 1, a structured fact is visually encoded by sharing convolutional layer parameters (denoted by $\theta_c$), and fully connected layer parameters (denoted by $\theta_u$); see Fig. 5(a). Then $W^S$, $W^P$, and $W^O$ transformation matrices are applied to produce $\mathbf{v}_S = \phi_S^{\mathcal{V}}(\mathbf{f}_v), \mathbf{v}_P = \phi_P^{\mathcal{V}}(\mathbf{f}_v)$ , and $\mathbf{v}_O = \phi_O^{\mathcal{V}}(\mathbf{f}_v)$. If we define $b = \psi(\theta_u, \psi(\theta_c, \mathbf{f}_v))$, then

$$\mathbf{v}_S = \phi_S^{\mathcal{V}}(\mathbf{f}_v) = W^S\, b, \quad \mathbf{v}_P = \phi_P^{\mathcal{V}}(\mathbf{f}_v) = W^P\, b, \quad \mathbf{v}_O = \phi_O^{\mathcal{V}}(\mathbf{f}_v) = W^O\, b. \qquad (4)$$

**Model 2 (structured fact CNN image encoder):** In contrast to Model 1, we use different convolutional layers for $S$ than that for $P$ and $O$, inspired by the idea that $P$ and $O$ are modifiers to $S$ (Fig. 5(b)). Starting from $\mathbf{f}_v$, there is a common set of convolutional layers, denoted by $\theta_c^0$, then the network splits into two branches, producing two sets of convolutional layers $\theta_c^S$ and $\theta_c^{PO}$, followed by two sets of fully connected layers $\theta_u^S$ and $\theta_u^{PO}$. If we define the output of the common S,P,O layers as $d = \psi(\theta_c^0, \mathbf{f}_v)$ and the output of the P,O column as $e = \psi(\theta_u^{PO}, \psi(\theta_c^{PO}, d))$, then

$$\mathbf{v}_S = \phi_S^{\mathcal{V}}(\mathbf{f}_v) = W^S \ \psi(\theta_u^S, \psi(\theta_c^S, d)), \ \mathbf{v}_P = \phi_P^{\mathcal{V}}(\mathbf{f}_v) = W^P \ e, \ \mathbf{v}_O = \phi_O^{\mathcal{V}}(\mathbf{f}_v) = W^O \ e. \tag{5}$$

**Structured fact RNN language encoder:** The structured fact language view is encoded using RNN word embedding vectors for $S$, $P$ and, $O$ separately. Hence

$$\mathbf{l}_S = \phi_S^{\mathcal{L}}(\mathbf{f}_l) = \mathrm{RNN}_{\theta_l}(\mathbf{f}_l^S), \mathbf{l}_P = \phi_P^{\mathcal{L}}(\mathbf{f}_l) = \mathrm{RNN}_{\theta_l}(\mathbf{f}_l^P), \mathbf{l}_O = \phi_O^{\mathcal{L}}(\mathbf{f}_l) = \mathrm{RNN}_{\theta_l}(\mathbf{f}_l^O) \tag{6}$$

where $\mathbf{f}_l^S$, $\mathbf{f}_l^P$, and $\mathbf{f}_l^O$ are the Subject, Predicate, and Object parts of $\mathbf{f}_l \in \mathcal{L}$. For each of them, the literals are dropped. In our experiments, $\theta_l$ is fixed to a pre-trained word vector embedding model (e.g. [26,31]) for $\mathbf{f}_l^S$, $\mathbf{f}_l^P$, and $\mathbf{f}_l^O$; see Fig 5(c).

**Loss function:** One way to model $p(\mathbf{f}_v, \mathbf{f}_l)$ for Model 1 and Model 2 is to assume that $p(\mathbf{f}_v, \mathbf{f}_l) \propto= \exp(-\mathrm{loss}_w(\mathbf{f}_v, \mathbf{f}_l))$ and minimize the distance $\mathrm{loss}_w(\mathbf{f}_v, \mathbf{f}_l)$ defined as

$$\mathrm{loss}_w(\mathbf{f}_v, \mathbf{f}_l) = w_S^{\mathbf{f}} \cdot D(\mathbf{v}_S, \mathbf{l}_S) + w_P^{\mathbf{f}} \cdot D(\mathbf{v}_P, \mathbf{l}_P) + w_O^{\mathbf{f}} \cdot D(\mathbf{v}_O, \mathbf{l}_O). \tag{7}$$

where $D(\cdot, \cdot)$ is a distance function. Thus we minimize the distance between the embeddings of the visual view and the language view. Our solution to penalize wild-card facts is to ignore their wild-card modifiers in the loss. Here $w_S^{\mathbf{f}} = 1$, $w_P^{\mathbf{f}} = 1$, $w_O^{\mathbf{f}} = 1$ for <S,P,O> facts , $w_S^{\mathbf{f}} = 1$, $w_P^{\mathbf{f}} = 1$, $w_O^{\mathbf{f}} = 0$ for <S,P> facts, and $w_S^{\mathbf{f}} = 1$, $w_P^{\mathbf{f}} = 0$, $w_O^{\mathbf{f}} = 0$ for <S> facts. Hence $\mathrm{loss}_w$ does not penalize the $O$ modifier for second-order facts or the $P$ and $O$ modifiers for first-order facts, which follows our definition of wild-cards. In this paper, we used $D(\cdot, \cdot)$ as the standard Euclidean distance.

**Testing (Two-view retrieval):** After training a model (either Model 1 or 2), we embed all the testing $\mathbf{f}_v$s (images) by the learnt models, and similarly embed all the test $\mathbf{f}_l$s as shown in Eq 6. For language view retrieval (retrieve relevant facts in language given an image), we compute the distance between the structured embedding of an image $\mathbf{v}$ and all the facts structured language embeddings $\mathbf{l}$s, which indicates relevance for each fact $\mathbf{f}_l$ for the given image. For visual view retrieval (retrieve relevant images given fact in language form), we compute the distance between the structured embedding of the given fact $\mathbf{l}$ and all structured visual embedding of images $\mathbf{v}$s in the test set. For first and second order facts, the wild-card part is ignored while computing the distance.

## 5  Experiments

### 5.1  Data Collection of Structured Facts

In order to train a model that connects the structured fact language view in $\mathcal{L}$ with its visual view in $\mathcal{V}$, we need to collect large scale data in the form of $(\mathbf{f}_v, \mathbf{f}_l)$ pairs. Large scale data collection is challenging in our setting since it relies on the localized association of a structured language fact $\mathbf{f}_l$ with an image $\mathbf{f}_v$ when such facts occur. In particular, it is a complex task to collect annotations for second-order facts and third-order facts.

We began our data collection by augmenting existing datasets with fact language view labels $\mathbf{f}_l$: PPMI [41], Stanford40 [42], Pascal Actions [11], Sports [16], Visual Phrases [34], INTERACT [4] datasets. The union of these 6 datasets resulted in 186 facts with 28,624 images as broken out in Table 1. We also extracted structured facts from the Scene Graph dataset [18] with 5000 manually annotated images in a graph structure from which first-, second-, and third-order relationships can be extracted. We extracted 110,000 second-order facts and 112,000 third-order facts. The majority of these are positional relationships. We also added to the aforementioned data, 380,000 second and third order fact annotation collected from MSCOCO and Flickr30K Entities datasets using a language approach as detailed in [27] in the supplementary. We show later in this section how we use this data to perform several experiments varying in scale to validate our claims. Table 2 shows the unique facts in the large scale dataset.

## 5.2 Setup of our Models and the designed Baselines

In our Model 1 and Model 2, $\theta_l$ is the GloVE840B RNN model [31] to encode structured facts in the language view.

1. **Model 1**: Model 1 is constructed from VGG-16, where $\theta_c$ is built from the layer conv_1_1 to pool5, and $\theta_u$ is the two following fully connected layers fc6 and fc7 in VGG-16 [36]. Similar to Model 2, $W^S$, $W^P$, and $W^O$ are initialized randomly and the rest of the parameters are initialized from VGG-16 trained on ImageNet [7].

2. **Model 2:** The shared layers $\theta_c^0$ match the architecture of the convolutional layers and pooling layer in VGG-16 named conv_1_1 until pool3, and have seven convolution layers. The subject layers $\theta_c^S$ and predicate-object layers $\theta_c^{PO}$ are two branches of convolution and pooling layers with the same architecture as VGG-16 layers named conv_4_1 until pool5 layer, which makes six convolution-pooling layers in each branch. Finally, $\theta_u^S$ and $\theta_u^{PO}$ are two instances of fc6 and fc7 layers in VGG-16 network. $W^S$, $W^P$, and $W^O$ are initialized randomly and the rest are initialized from VGG-16 trained on ImageNet.

3. **Multiview CCA IJCV14 [15] (MV CCA)** : MV CCA expects features from both views. For visual view features, we used VGG16 (FC6). For the language view features, we used GloVE. Since MV CCA does not support wild-cards, we fill the wild-card parts of $\Phi^{\mathcal{L}}(\mathbf{f}_l)$ with zeros for First Order and Second order facts.

4. **ESZSL ICML15 Baseline [33] (ESZSL)**: ESZSL also expects both visual and semantic features for a fact. As in MV CCA, we used VGG16 (FC6) and GloVE.

Table 1: Our fact augmentation of six datasets

|  | Unique language views $\mathbf{f}_l$ | | | | Number of ( $\mathbf{f}_v$, $\mathbf{f}_l$) pairs | | | |
|---|---|---|---|---|---|---|---|---|
|  | S . | SP. | SPO . | total | S | SP | SPO | total images |
| INTERACT | 0 | 0 | 60 | 60 | 0 | 0 | 3171 | 3171 |
| VisualPhrases | 11 | 4 | 17 | 32 | 3594 | 372 | 1745 | 5711 |
| Stanford40 | 0 | 11 | 29 | 40 | 0 | 2886 | 6646 | 9532 |
| PPMI | 0 | 0 | 24 | 24 | 0 | 0 | 4209 | 4209 |
| SPORT | 14 | 0 | 6 | 20 | 398 | 0 | 300 | 698 |
| Pascal Actions | 0 | 5 | 5 | 10 | 0 | 2640 | 2663 | 5303 |
| Union | 25 | 20 | 141 | 186 | 3992 | 5898 | 18734 | 28624 |

Table 2: Large Scale Dataset

|  | S | SP | SPO | Total |
|---|---|---|---|---|
| Training facts | 6116 | 57681 | 107472 | 171269 |
| Testing facts | 2733 | 22237 | 33447 | 58417 |
| Train/Test Intersection | 1923 | 13043 | 11774 | 26740 |
| Test unseen facts | 810 | 9194 | 21673 | 31677 |

5. **Image-Sentence Similarity (TACL15 [21]) (MS COCO pretrained)**: We used the theano implementations of this method that were made publically available by the authors [20]. The purpose of applying MS COCO pretrained image-caption models is to show how image-caption trained models perform when applied to fact level recognition in our setting. In order to use these models to measure simility between image and facts in our setting, we provide them with the image and a phrase constructed from the fact language representation. For example <person, riding, horse > is converted to "person riding horse".

6. **Image-Sentence Similarity (TACL15 [21] (retrained)**: In contrast to the previous setting, we retrain these models by providing them our image-fact training pairs where facts are converted to phrases. The results show the value of learning models on the fact level instead of the caption level.

### 5.3   Evaluation Metrics

We present evaluation metrics for both language view retrieval and visual view retrieval.
**Metrics for visual view retrieval (retrieving $\mathbf{f}_v$ given $\mathbf{f}_l$):** To retrieve an image (visual view) given a language view (e.g. <S: person, P: riding, O: horse>), we measure the performance by mAP (Mean Average Precision).An image $\mathbf{f}_v$ is considered positive only if there is a pair $(\mathbf{f}_l, \mathbf{f}_v)$ in the annotations. Even if the retrieved image is relevant but such pair does not exist, it is considered incorrect. We also use mAP10, mAP100 variants that compute the mAP based on only the top 10 or 100 retrieved images, which is useful for evaluating large scale experiments.
**Metrics for language view retrieval (retrieving $\mathbf{f}_l$ given $\mathbf{f}_v$):** To retrieve fact language views given an image. we use top 1, top 5, top 10 accuracy for evaluation. We also used MRR (mean reciprocal ranking) metric which is basically $1/r$ where $r$ is the rank of the correct class. An important issue with our setting is that there might be multiple facts in the same image. Given that there are $L$ correct facts in the given image to achieve top 1 performance these $L$ facts must all be in the top $L$ retrieved facts. Accordingly, top K means the $L$ facts are in the top $L + K - 1$ retrieved facts. A fact language view $\mathbf{f}_l$ is considered correct only if there is a pair $(\mathbf{f}_l, \mathbf{f}_v)$ in the annotations.

It is not hard to see that the aforementioned metrics are very harsh, especially in the large scale setting. For instance, if the correct fact for an image is <S:man,P: jumping>, then an answer <S:person, P:jumping> receives zero credit. Also, the evaluation is limited to the ground truth fact annotations. There might be several facts in an image but the provided annotations may miss some facts. Qualitatively we found the metrics harsh for our large scale experiment. Defining better metrics is future work.

### 5.4   Small and Mid scale Experiments

We performed experiments on several datasets ranging in scale: Stanford40 [42], Pascal Actions [41], Visual Phrases [34], and the union of six datasets described earlier in Table 1 in Sec. 5.1. We used the training and test splits defined with those datasets. For the union of six datasets, we unioned the training and testing annotations to get the final split. In all these training/testing splits, each fact language view $\mathbf{f}_l$ has corresponding tens of visual views $\mathbf{f}_v$ (i.e., images) split into training and test sets. So, each test image belongs to a fact that is seen by other images in the training set.

Table 3 shows the performance of our Model 1, Model 2, and the designed baselines on these four datasets for both view retrieval tasks. We note that Model 2 works relatively better than Model 1 as the scale size increases as shown here when comparing results on Pascal dataset to larger datasets like Stanford40, Visual Phrases, and 6DS. In the next section, we show that Model2 is clearly better than Model 1 in the large scale setting. Our intuition behind this result is that Model 2 learns a different set of convolutional filters in the PO branch to understand action/attributes and interactions which is different from the filter bank learned to discriminate between different subjects for the S branch. In contrast, Model 1 is trained by optimizing one bank of filters for SPO altogether, which might conflict to optimize for both S and PO together; see Fig 5.

Learning from image-caption pairs even on big dataset like MSCOCO does not help discriminate between tens of facts as shown in these experiments. However, retraining these models by providing them image-fact pairs makes them perform much better as shown in Table 3. Compared to other methods on language view retrieval, we found Model 1 and 2 perform significantly better than TACL15 [21] even when retrained for our setting, especially on PASCAL10, Stanford40, and 6DS datasets which are dominated by SP and SPO facts; see Table 1. For visual view retrieval, performance is competitive in some of the datasets. We think the reason is due to the structure that makes our models relate all fact types by the visual modifiers notion.

Although ESZSL is applicable in our setting, it is among the worst performing methods in Table 3. This could be because ESZSL is mainly designed for Zero-Shot Learning, but each fact has some training examples in these experiments. Interestingly, MV CCA with the chosen visual and language features is among the best methods. Next we compare these methods when number of facts becomes three orders of magnitudes larger and with tens of thousands of testing facts that are unseen in training.

Table 3: Small and Medium Scale Experiments

| | | Language View retrieval | | | Visual View retrieval | | |
|---|---|---|---|---|---|---|---|
| | | Top1 | Top 5 | MRR | mAP | mAP10 | mAP100 |
| **Standord40 (40 facts)** | **Model2** | **74.46** | **92.01** | **82.26** | 73 | 98.35 | 92 |
| (11 SP, 29 SPO) | **Model1** | 71.22 | 90.98 | 82.09 | **74.57** | **99.72** | **92.62** |
| | **MV CCA IJCV14** | 67.74 | 88.32 | 76.80 | 66.00 | 96.86 | 86.66 |
| | **ESZSL ICML15 [33]** | 40.89 | 74.93 | 56.08 | 50.9 | 93.87 | 78.35 |
| | **Image-Sentence TACL15 [21] (COCO pretrained)** | 33.73 | 62.62 | 47.70 | 26.29 | 59.68 | 44.2 |
| | **Image-Sentence TACL15 [21] (retrained)** | 60.86 | 87.82 | 72.51 | 51.9 | 88.13 | 74.55 |
| | **Chance** | 2.5 | - | - | - | - | - |
| **Pascal Actions (10 facts)** | **Model2** | **74.760** | **95.750** | **83.680** | **80.950** | **100.000** | **97.240** |
| (5 SP, 5 SPO) | **Model1** | 74.080 | 95.790 | 83.280 | 80.530 | 100.000 | 96.960 |
| | **MV CCA IJCV14** | 59.82 | 92.78 | 73.16 | 33.45 | 66.52 | 53.29 |
| | **ESZSL ICML15 [33]** | 44.846 | 88.864 | 63.366 | 54.274 | 89.968 | 82.273 |
| | **Image-Sentence TACL15 [21] (COCO pretrained)** | 46.050 | 86.907 | 62.796 | 40.712 | 88.694 | 71.078 |
| | **Image-Sentence TACL15 [21] (retrained)** | 60.27 | 94.66 | 74.77 | 50.58 | 84.65 | 71.61 |
| | **Chance** | 10 | - | - | - | - | - |
| **VisualPhrases (31 facts)** | **Model2** | **34.367** | **76.056** | **47.263** | **39.865** | **61.990** | **48.246** |
| (14 S, 4 SP, 17 SPO) | **Model1** | 28.100 | 75.285 | 42.534 | 38.326 | 65.458 | 46.882 |
| | **MV CCA IJCV14 [15]** | 28.94 | 70.61 | 88.92 | 28.27 | 49.30 | 34.48 |
| | **ESZSL ICML15 [33]** | 33.830 | 68.264 | 44.650 | 33.010 | 57.861 | 41.131 |
| | **Image-Sentence TACL15 [21] (COCO pretrained)** | 30.111 | 64.494 | 42.777 | 26.941 | 49.892 | 33.014 |
| | **Image-Sentence TACL15 [21] (retrained)** | 32.32 | 94.72 | 50.7 | 28.0 | 49.89 | 33.21 |
| | **Chance** | 3.2 | - | - | - | - | - |
| **6DS (186 facts)** | **Model2** | **69.63** | **80.32** | **70.66** | **34.86** | **61.03** | **50.68** |
| (25 S, 20 SP, 141 SPO) | **Model1** | 68.94 | 78.74 | 70.74 | 34.64 | 56.54 | 47.87 |
| | **MV CCA IJCV14 [15]** | 29.84 | 39.78 | 32.00 | 23.93 | 46.43 | 36.44 |
| | **ESZSL ICML15 [33]** | 27.53 | 47.4 | 58.2 | 30.7 | 60.97 | 47.58 |
| | **Image-Sentence TACL15 [21] (COCO pretrained)** | 15.71 | 26.84 | 19.65 | 9.37 | 21.58 | 15.88 |
| | **Image-Sentence TACL15 [21] (retrained)** | 26.13 | 41.10 | 30.94 | 26.17 | 56.10 | 40.4 |
| | **Chance** | 0.54 | - | - | - | - | - |

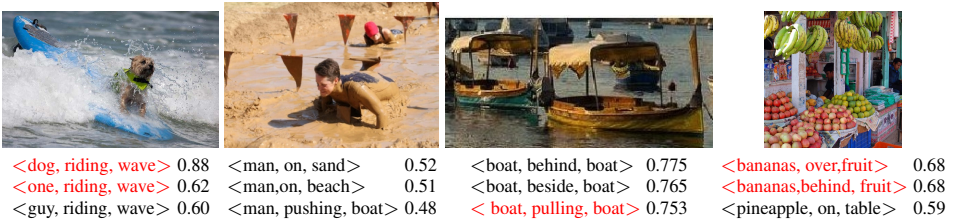

<dog, riding, wave> 0.88    <man, on, sand>      0.52    <boat, behind, boat> 0.775    <bananas, over, fruit>      0.68
<one, riding, wave> 0.62    <man, on, beach>     0.51    <boat, beside, boat> 0.765    <bananas, behind, fruit> 0.68
<guy, riding, wave> 0.60    <man, pushing, boat> 0.48    <boat, pulling, boat> 0.753    <pineapple, on, table>      0.59

Fig. 6: Language View Retrieval examples (red means unseen facts)

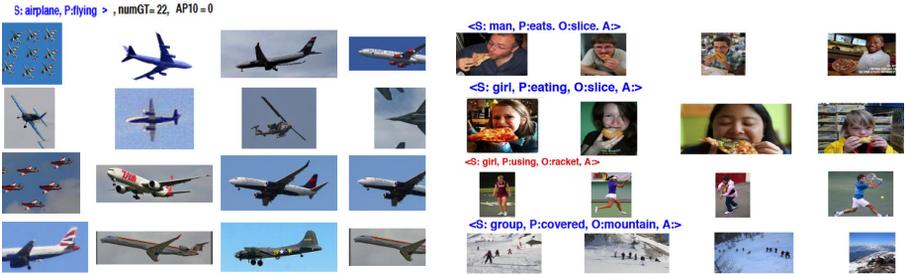

Fig. 7: Visual View Retrieval Examples (red means unseen facts)

## 5.5   Large Scale Experiment

In this experiment, we used the union of all the data described in Sec. 5.1. We further augmented this data with 2000 images for each MS COCO object (80 classes) as first-order facts. We also used object annotations in the Scene Graph dataset as first-order fact annotations with a maximum of 2000 images per object. Finally, we randomly split all the annotations into an 80%-20% split, constructing sets of 647,746 $(\mathbf{f}_v, \mathbf{f}_l)$ training pairs (with 171,269 unique fact language views $\mathbf{f}_l$) and 168,691 $(\mathbf{f}_v, \mathbf{f}_l)$ testing pairs (with 58,417 unique $\mathbf{f}_l$), for a total of $(\mathbf{f}_v, \mathbf{f}_l)$ 816,436 pairs, 202,946 unique $\mathbf{f}_l$. Table 2 shows the coverage of different types of facts. There are 31,677 language view test facts that were unseen in the training set (851 <S>, 9,194 <S,P>, 21,673 <S,P,O>). The majority of the facts have only one example; see the supplementary material.

Qualitative results are shown in Fig. 6, 7 (with many more in the supplementary). In Fig. 6, our model's ability to generalize can be seen in the red facts. For example, for the leftmost image our model was able to correctly identify the image as <dog, riding, wave> despite that fact never being seen in our training data. The left images in Fig. 7 show the variety of images we can retrieve for the query <airplane, flying>. In the right images in Fig. 7, note how our model learns to visually distinguish gender ("man" versus "girl"), and group versus single. It can also correctly retrieve images for facts that were never seen in the training set (<girl, using, racket>). Highlighting the harshness of the metric, Fig. 7 also shows that <airplane, flying> has zero AP10 value giving us zero credit since the top images were just annotated as an < airplane>.

To perform retrieval in both directions, we used the FLANN library [28] to compute the (approximate) 100 nearest neighbors for $\mathbf{f}_l$ given $\mathbf{f}_v$, and vice-versa. Details about the nearest-neighbor database creation and the large scale evaluation could be found in the supplementary. The results in Table 4 indicate that Model 2 is better than Model 1 for retrieval from both views, which is consistent with our medium scale results and our intuition. Model 2 is also multiple orders of magnitude better than chance and is also significantly better than the competing methods. To test the value of structure, we ran an experiment where we averaged the S, P, and O parts of the visual and language embedding vectors instead of keeping the structure. Removing the structure leads to a noticeable decrease in performance from 16.39% to 8.1% for the K1 metric; see Table 4.

Previous smaller scale experiments are orders of magnitudes smaller and also less challenging since all facts were seen during training. Figure 8 shows the effect of the scale on the Top1 performance for language view retrieval task (denoted K1). There is an observable increase on the improvement of Model 2 compared to the baselines in the large scale setting. Additionally, the performance of the image-caption similarity methods degrade substantially. We think this is due to both the large scale of the facts and that the majority of the facts have zero or very few training examples. Interestingly, MV CCA is among the best performing methods in the large scale setting. However, Model 2 and Model 1 outperform MV CCA on both Top1 and Top 5 metrics; see Table 4. On the language view retrieval, we have very competitive results to MV CCA but as we have notices several good visual retrieval results for which the metric gives zero-credit.

Figure 13 shows the Top10 large scale knowledge view retrieval (K10) results reported in Table 4 broken out by fact type and the number of images per fact. These results show that Model 2 generally behaves better with compared other models with the increase of facts. We noticed a slight increase for Model 1 over Model 2.

It is desirable for a method to be able to generalize to understand an SPO interaction from training examples involving its components, even when there are zero or very few training examples for the exact SPO with all its parts S,P and O. Table 5 shows the K10 performance for SPOs where the number of training examples is $\leq 5$. For example, the column **SP≥15, O≥15** means $\leq 5$ examples of an SPO that has at least 15 examples for the SP part and for the O part. An example of this case is when we see zero or very few examples of <person, petting, horse>, but we see at least 15 examples of <person, petting, something=dog/cat/etc (not horse)> and at least 15 examples of something interacting with a horse <*,*, horse>. Model2 performs the best in all the listed generalization cases in Table 5. We found a similar generalization behavior for SP facts that have no more than 5 examples during training. We add more figures and additional results in the supplementary materials.

Table 4: Large Scale Experiment

| | Language View retrieval % | | | Visual view Retrieval % | |
|---|---|---|---|---|---|
| | Top1 | Top 5 | Top 10 | mAP100 | mAP10 |
| **Model 2** | **16.39** | **17.62** | **18.41** | 0.90 | 0.90 |
| **Model 1** | 13.27 | 14.19 | 14.80 | 0.73 | 0.73 |
| **Model 2 (Unstructured by SPO average)** | 8.1 | 12.4 | 14.00 | 0.61 | 0.62 |
| **MV CCA IJCV14 [15]** | 12.28 | 12.84 | 13.15 | **1.0** | **1.0** |
| **ESZSL ICML15 [33]** | 5.80 | 5.84 | 5.86 | 0.4 | 0.4 |
| **Image-Sentence TACL15 [21] (COCO pretrained)** | 3.48 | 3.48 | 3.5 | 0.021 | 0.0087 |
| **Image-Sentence TACL15 [21] (retrained)** | 5.87 | 6.06 | 6.15 | 0.29 | 0.29 |
| **Chance** | 0.0017 | - | - | - | - |

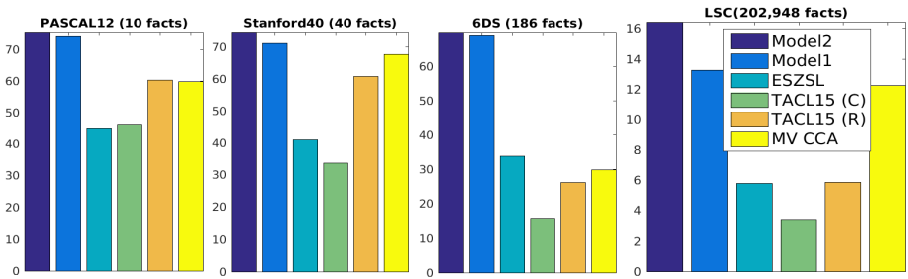

Fig. 8: K1 Performance Across Different Datasets. These graphs show the advantage of the proposed models as the scale increases from left to right. (R) for TACL15 means the retrained version, (C) means COCO pretrained model; see Sec 5.2

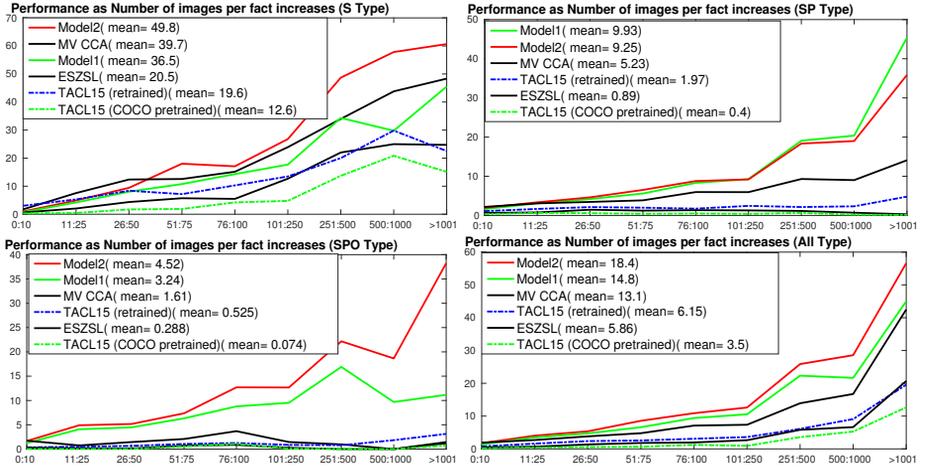

Fig. 9: K10 Performance ($y$-axis) versus the number of images per fact ($x$-axis). Top Left: Objects (S), Top Right: Attributed Objects and Objects performing Actions (SP), Bottom Left: Interactions (SPO), Bottom Right: All Facts.

Table 5: Generalization: SPO Facts of less than or equal 5 examples (K10 metric)

| Cases | SP≥15, O≥15 | PO≥15, ≥15 | SO≥15, P≥15 | S≥15, PO≥15 | SO≥15, PO≥15 | SO≥15, SP≥15 | S,P,O≥15 | S,P,O≥100 |
|---|---|---|---|---|---|---|---|---|
| NumFacts for this case | 10605 | 9313 | 4842 | 4673 | 1755 | 3133 | 21616 | 12337 |
| Model2 | **2.063** | **2.026** | **3.022** | **2.172** | **3.092** | **2.962** | **1.861** | **2.462** |
| Model1 | 1.751 | 1.357 | 1.961 | 1.645 | 1.684 | 2.097 | 1.405 | 1.666 |
| ESZSL | 0.149 | 0.107 | 0.098 | 0.066 | 0.041 | 0.038 | 0.240 | 0.176 |
| TACL15 (COCO pretrained) | 0.013 | 0.024 | 0.025 | 0.019 | 0.000 | 0.013 | 0.034 | 0.027 |
| TACL15 (retrained) | 0.367 | 0.380 | 0.473 | 0.384 | 0.543 | 0.586 | 0.353 | 0.438 |
| MV CCA | 1.221 | 1.889 | 1.462 | 1.273 | 1.786 | 1.109 | 1.853 | 1.838 |

## 6   Conclusion

We introduce new setting for learning unbounded number of facts in images, which could be referred to as a model for gaining visual knowledge. The facts could be of different types like objects, attributes, actions, and interactions. While studying this task, we consider Uniformity, Generalization, Scalability, Bi-directionality, and Structure. We investigated several baselines from multi-view learning literature which were adapted to the proposed setting. We proposed learning representation methods that outperform the designed baseline mainly by the advantage of relating facts by structure.

# References

1. Akata, Z., Perronnin, F., Harchaoui, Z., Schmid, C.: Label-embedding for attribute-based classification. In: Proceedings of the IEEE Conference on Computer Vision and Pattern Recognition. pp. 819–826 (2013)

2. Antol, S., Agrawal, A., Lu, J., Mitchell, M., Batra, D., Lawrence Zitnick, C., Parikh, D.: Vqa: Visual question answering. In: ICCV (2015)

3. Antol, S., Zitnick, C.L., Parikh, D.: Zero-Shot Learning via Visual Abstraction. In: ECCV (2014)

4. Antol, S., Zitnick, C.L., Parikh, D.: Zero-shot learning via visual abstraction. In: ECCV (2014)

5. Ba, J., Swersky, K., Fidler, S., Salakhutdinov, R.: Predicting deep zero-shot convolutional neural networks using textual descriptions. In: ICCV (2015)

6. Chen, C.Y., Grauman, K.: Inferring analogous attributes. In: CVPR (2014)

7. Deng, J., Dong, W., Socher, R., Li, L.J., Li, K., Fei-Fei, L.: Imagenet: A large-scale hierarchical image database. In: CVPR. IEEE (2009)

8. Devlin, J., Cheng, H., Fang, H., Gupta, S., Deng, L., He, X., Zweig, G., Mitchell, M.: Language models for image captioning: The quirks and what works. arXiv preprint arXiv:1505.01809 (2015)

9. Devlin, J., Gupta, S., Girshick, R., Mitchell, M., Zitnick, C.L.: Exploring nearest neighbor approaches for image captioning. arXiv preprint arXiv:1505.04467 (2015)

10. Elhoseiny, M., Saleh, B., Elgammal, A.: Write a classifier: Zero-shot learning using purely textual descriptions. In: ICCV (2013)

11. Everingham, M., Van Gool, L., Williams, C.K.I., Winn, J., Zisserman, A.: The PASCAL Visual Object Classes Challenge 2012 (VOC2012) Results. http://www.pascal-network.org/challenges/VOC/voc2012/workshop/index.html

12. Farhadi, A., Endres, I., Hoiem, D., Forsyth, D.: Describing objects by their attributes. In: CVPR (2009)

13. Frome, A., Corrado, G.S., Shlens, J., Bengio, S., Dean, J., Mikolov, T., et al.: Devise: A deep visual-semantic embedding model. In: NIPS (2013)

14. Gkioxari, G., Malik, J.: Finding action tubes. In: CVPR (2015)

15. Gong, Y., Ke, Q., Isard, M., Lazebnik, S.: A multi-view embedding space for modeling internet images, tags, and their semantics. International journal of computer vision 106(2), 210–233 (2014)

16. Gupta, A.: Sports Dataset. http://www.cs.cmu.edu/~abhinavg/Downloads.html (2009), [Online; accessed 15-July-2015]

17. Jia, Y., Shelhamer, E., Donahue, J., Karayev, S., Long, J., Girshick, R., Guadarrama, S., Darrell, T.: Caffe: Convolutional architecture for fast feature embedding. In: Proceedings of the ACM International Conference on Multimedia. pp. 675–678. ACM (2014)

18. Johnson, J., Krishna, R., Stark, M., Li, L.J., Shamma, D., Bernstein, M., Fei-Fei, L.: Image retrieval using scene graphs. In: CVPR (2015)

19. Karpathy, A., Joulin, A., Li, F.F.F.: Deep fragment embeddings for bidirectional image sentence mapping. In: Advances in neural information processing systems. pp. 1889–1897 (2014)

20. Kiros, J.R.: Image-sentence tacl15 implementation. https://github.com/ryankiros/visual-semantic-embedding (2015), [Online; accessed 19-Nov-2015]

21. Kiros, R., Salakhutdinov, R., Zemel, R.S.: Unifying visual-semantic embeddings with multimodal neural language models. TACL (2015)

22. Krizhevsky, A., Sutskever, I., Hinton, G.E.: Imagenet classification with deep convolutional neural networks. In: NIPS (2012)
23. Lampert, C.H., Nickisch, H., Harmeling, S.: Learning to detect unseen object classes by between-class attribute transfer. In: CVPR (2009)
24. Malinowski, M., Rohrbach, M., Fritz, M.: Ask your neurons: A neural-based approach to answering questions about images. In: ICCV (2015)
25. Mao, J., Xu, W., Yang, Y., Wang, J., Yuille, A.: Deep captioning with multimodal recurrent neural networks (m-rnn). ICLR (2015)
26. Mikolov, T., Sutskever, I., Chen, K., Corrado, G.S., Dean, J.: Distributed representations of words and phrases and their compositionality. In: NIPS (2013)
27. Mohamed Elhoseiny, Scott Cohen, W.C.B.P.A.E.: Automatic annotation of structured facts in images. In: Arxiv (2016)
28. Muja, M., Lowe, D.: Flann-fast library for approximate nearest neighbors user manual. Computer Science Department, University of British Columbia, Vancouver, BC, Canada (2009)
29. Norouzi, M., Mikolov, T., Bengio, S., Singer, Y., Shlens, J., Frome, A., Corrado, G.S., Dean, J.: Zero-shot learning by convex combination of semantic embeddings. In: ICLR (2014)
30. Palatucci, M., Pomerleau, D., Hinton, G.E., Mitchell, T.M.: Zero-shot learning with semantic output codes. In: NIPS (2009)
31. Pennington, J., Socher, R., Manning, C.D.: Glove: Global vectors for word representation. EMNLP (2014)
32. Ren, M., Kiros, R., Zemel, R.: Exploring models and data for image question answering. In: NIPS (2015)
33. Romera-Paredes, B., Torr, P.: An embarrassingly simple approach to zero-shot learning. In: Proceedings of The 32nd International Conference on Machine Learning. pp. 2152–2161 (2015)
34. Sadeghi, M.A., Farhadi, A.: Recognition using visual phrases. In: CVPR (2011)
35. Salakhutdinov, R., Torralba, A., Tenenbaum, J.: Learning to share visual appearance for multiclass object detection. In: Computer Vision and Pattern Recognition (CVPR), 2011 IEEE Conference on. pp. 1481–1488. IEEE (2011)
36. Simonyan, K., Zisserman, A.: Very deep convolutional networks for large-scale image recognition. In: ICLR (2015)
37. Socher, R., Ganjoo, M., Sridhar, H., Bastani, O., Manning, C.D., Ng, A.Y.: Zero shot learning through cross-modal transfer. In: NIPS (2013)
38. Szegedy, C., Liu, W., Jia, Y., Sermanet, P., Reed, S., Anguelov, D., Erhan, D., Vanhoucke, V., Rabinovich, A.: Going deeper with convolutions (June 2015)
39. Vinyals, O., Toshev, A., Bengio, S., Erhan, D.: Show and tell: A neural image caption generator (2015)
40. Xu, K., Ba, J., Kiros, R., Courville, A., Salakhutdinov, R., Zemel, R., Bengio, Y.: Show, attend and tell: Neural image caption generation with visual attention. In: ICML (2015)
41. Yao, B., Fei-Fei, L.: Grouplet: A structured image representation for recognizing human and object interactions. In: CVPR (2010)
42. Yao, B., Jiang, X., Khosla, A., Lin, A.L., Guibas, L., Fei-Fei, L.: Human action recognition by learning bases of action attributes and parts. In: ICCV (2011)
43. Zhang, N., Paluri, M., Ranzato, M., Darrell, T., Bourdev, L.: Panda: Pose aligned networks for deep attribute modeling. In: Computer Vision and Pattern Recognition (CVPR), 2014 IEEE Conference on. pp. 1637–1644. IEEE (2014)
44. Zhou, B., Lapedriza, A., Xiao, J., Torralba, A., Oliva, A.: Learning deep features for scene recognition using places database. In: NIPS (2014)
45. Zipf, G.K.: The psycho-biology of language. (1935)

## Supplementary Materials

The supplementary include the following materials

## 7   Large Scale Experiment Data

### 7.1   Training images of the test facts (related to Fig 9 in the paper)

Fig. 10 shows more details about the number of training image for each test fact. The $x-$ axis shows different ranges for each. The $y-$ axis shows the number of test facts whose number of training examples falls between the specified range of examples.

Note that the $x-$ axis here is the same as the $x-$ axis in Fig 9 in the paper.

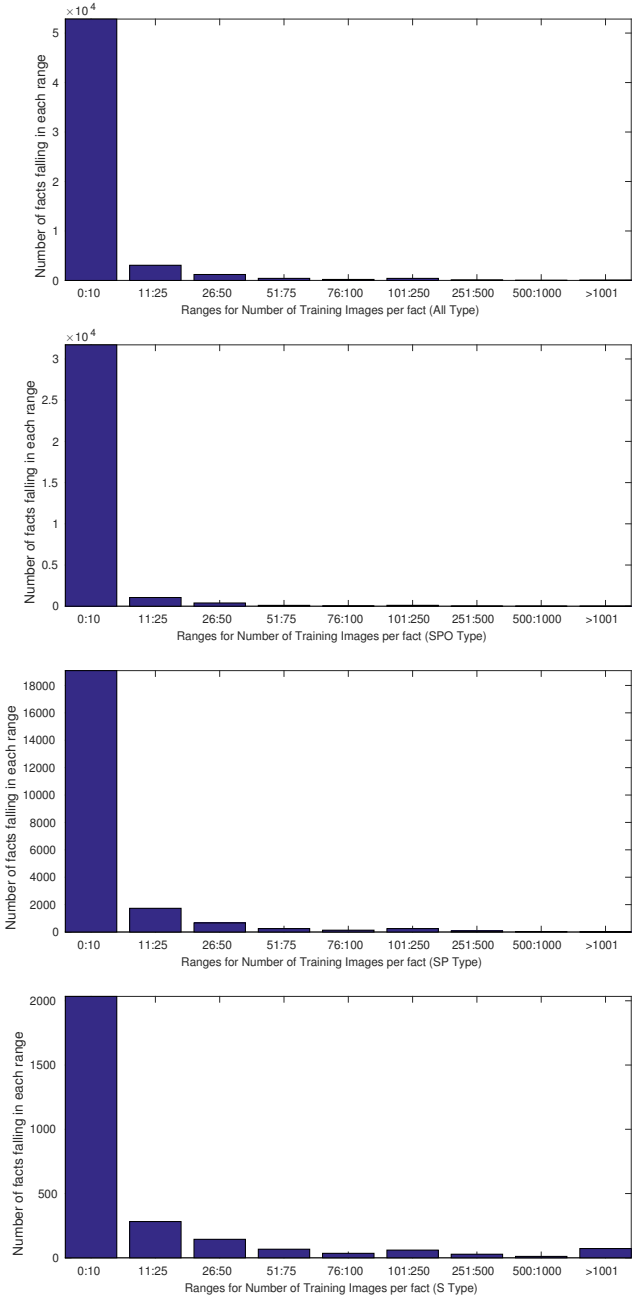

Fig. 10: Number of Facts per each "Number of Images Range". $x$−axis shows ranges of number of images per fact. $y$− axis is the number of facts whose number of images fall in the corresponding Range

### 7.2  Test images

We show the test examples in two groups.

- The first group is the group of facts that has at least one training image (seen fact) in Fig 11 .
- The second group of facts is the group where there is no training images at all; see Fig. 12.
- The $x-$axis shows the "fact identifier" where facts are sorted from largest to smallest number of images.
- $y-$axis the number of test images in each of them. Each figure has three plot, one for each fact type $<S\geq$, $<S,P\geq$, $<S,P,O\geq$.
- The figures show that in both cases, the majority of the facts have at most one test example.
- This quantitatively shows the difficulty of the evaluation especially for the large scale setting for those facts.

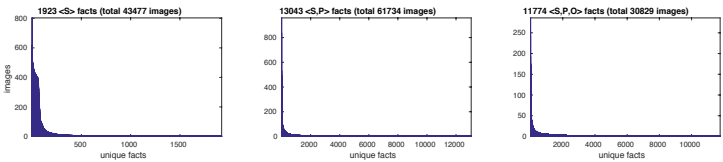

Fig. 11: 26,740 unique test facts that have at least one training example (seen facts), total of 136,040 images ($x-$ axis shows these facts, $y-$ axis shows the number of test images per each fact)

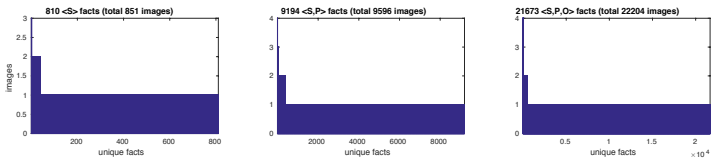

Fig. 12: 31,677 unique unseen test facts, total of 32,651 images ($x-$ axis shows these facts, $y-$ axis shows the number of test images per each fact)

## 8    Training and Implementation Details

**GPU Framework** Our implementation was based on Caffe [17] for our implementation.

**Training Parameters** Model 1 and Model2 were trained by back propagation with stochastic gradient descent. The base learning rate is assigned $0.5 \times 10^{-4}$. For fine-tuning, the learning rate of the randomly initialized parameters are assigned to be ten times faster than the learning rate of the remaining parameters( initialized from the pretrained CNN). Please refer to the paper where randomly initialized parameters are specified for each of Model 1 and Model 2.

The decay of the learning rate $\gamma$ is 0.1. While training our CNNs, we drop the learning rate by a factor of $\gamma$ every 5000 iterations. The momentum and the weight decay were assigned to 0.9 and 0.0001 respectively. Training images are randomly shuffled before feeding the CNN for training. The training batch size was 100 images.

**Training and testing batches** At training time, we randomly sample 224x224 batches from the down-scaled 256x256 images. At test time the center 224x224 batches are taken.

**Normalization of GloVE word vectors [31]:** For the structured fact language encoder (Fig 4 c in the paper), we normalize the S, P, and O vectors to L2 norm 1 individually. Then we subtract the mean of all the training vectors. This is similar to subtracting the mean of the image for encoding the visual view of the image. For first-order, we fill the P and O parts with zeros. For second-order facts, we fill the O part with zeros.

As illustrated in the paper, Model 1 and Model 2 do not penalize first order facts for P and O, and do not penalize second order facts for O; see Eq 7 in the paper (lines 333 to 334, page 8).

## 9   Details about the Language View Retrieval Metric in the Large Scale Experiment used in our experiments

In the language view retrieval metric used in our submission, we created a database for all facts of 900 dimensional vectors (300 dimensions for S, followed by 300 dimensions for P, followed by 300 dimensions for O). It may not be hard to see that, our methods leans to produce more specific facts ( higher order facts compared to lower order facts; highest order fact is the third order facts). This is because lower order facts have zeros in the unspecified fact components.

In order to avoid incorrectly penalize a method for being more specific, we do not penalize more specific facts of the ground truth facts. This cases only happens for ground truth first and second-order facts.

For first-order ground truth facts $<S>$, the retrieved second $<S,P>$ and third order facts $<S,P,O>$ that have exactly the same ground truth subject S are not penalized. For example if the ground truth fact is $<car>$ and the retrieved fact is $<car, red>$.

For second-order ground $<S,P>$ truth facts, the retrieved third order facts that have exactly the same ground truth subject S and predicate P are not penalized. For example if the ground truth is $<person, playing>$ and the retrieved fact is $<person, playing, guitar>$.

We attach the code that performs the evaluation in our experiments from which our results could be reproduced. It could also be used to evaluate any other method for comparison to our work. We name this metric as "Metric 1"

**Attachment:** The implementation of Metric 1 could be found in the attached **Get-SherlockResults_ANN_metric1.m**.

We also report all the results that perform tagging for each fact type separately in Sec 10 in this supplementary.

# 10    Language Retrieval Result with Metric2 Defined below ( uses a KD Tree database for Each Fact Type)

In this section, we present more results, where we deal with each fact order separately at test time. There is no change during training; it is exactly the same model. At test time, we build three KD Tree databases, one for each fact order; i.e., one for first-order, one for second-order, and one for third order facts. At test time, we are given a test image that might include first-,second-, and/or third-order facts. We then use the given model to check the closest 100 facts in each database depending on its type. So, this step produces three lists, which are (1) relevant first-order facts,(2) second-order facts, (3) third-order facts given the image. Then, the rank for each ground truth facts is checked in its corresponding list based on its type.

For instance, if an image has two facts (such as $<$ man, riding, horse $>$, $<$man, Asian$>$)). We look-up the rank for $<$ man, riding, horse $>$ in the list of third-order facts and we look-up the rank for $<$man, Asian$>$ in the list of second-order facts.

In order to compute top K-performance, the rank of the ground truth facts are checked. Accordingly, top K means the $L$ facts are in the top $L + K - 1$ retrieved facts, where $L$ is the number of ground truth facts in the image of the same order. However, it is checked for each fact type separately for Metric 2 .

**This metric tests accuracy for a given order of specificity (i.e. which order fact are you interested in to describe a given image). We name this metric as "Metric 2".** The metric in the paper has a bias toward tagging with the most specific fact (third order). We name this metric in the main paper (pages 13 and 14) as **"Metric 1"**.

Accordingly, we can produce here a new set of results based on **"Metric 2"** as opposed to **"Metric 1"** which is reported in the paper (pages 13 and 14).

**Attachment:** The implementation of Metric 2 could be found in the attached **GetSherlockResults_ANN_metric2.m**.

## 10.1    Language View Retrieval Summary Table (Metric 2)

We see in Table  6 that our Model 2 outperforms all other methods in the large scale experiment using Metric 2, as it did for Metric 1 in the main paper.

Table 6: (Metric 2) Language View Retrieval Summary Results

|                          | K1     | K5     | K10    | MRR    |
|--------------------------|--------|--------|--------|--------|
| **Model2**               | 11.200 | 17.800 | 20.300 | 14.200 |
| **Model1**               | 9.220  | 15.400 | 18.000 | 12.000 |
| **MV CCA**               | 3.690  | 4.970  | 5.400  | 4.250  |
| **ESZSL**                | 2.350  | 3.070  | 3.300  | 2.680  |
| **TACL15 ( retrained )** | 0.870  | 1.650  | 1.940  | 1.190  |
| **TACL15 ( Coco pretrained )** | 0.080 | 0.169 | 0.282 | 0.153 |

## 10.2  Results as Number of Images per Fact Increases (Metric 2)

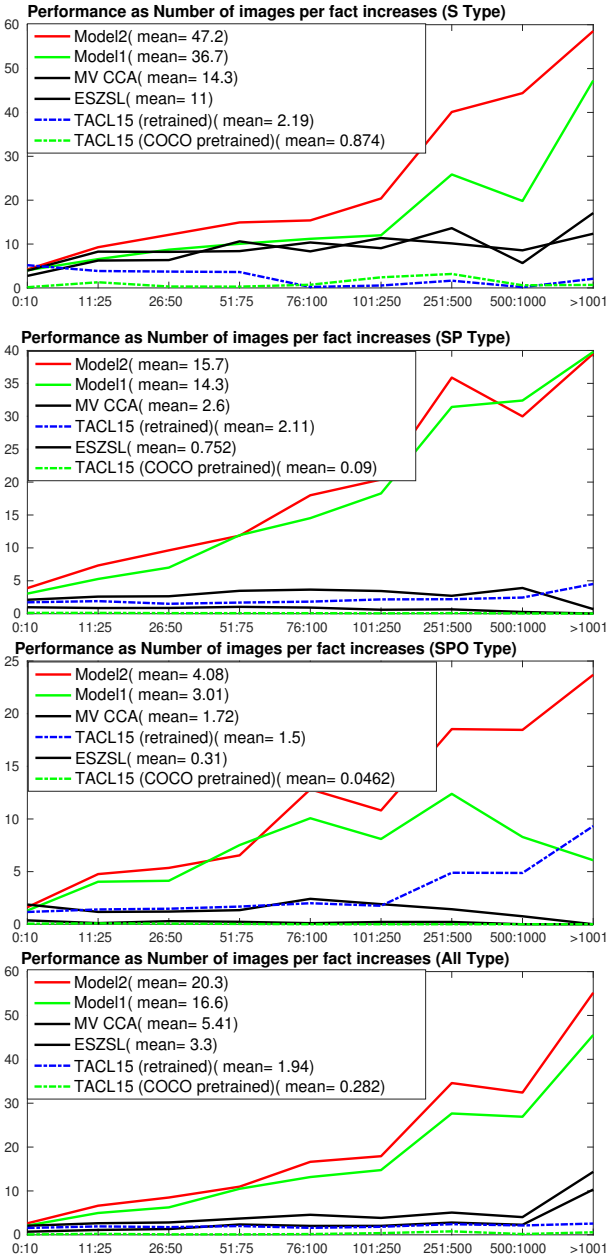

Fig. 13: (Metric 2) K10 Performance ($y$-axis) versus the number of images per fact ($x$-axis). First from Top: Objects (S), Second from Top: Attributed Objects and Objects performing Actions (SP), Third from Top: Interactions (SPO), Fourth from Top: All Facts.

## 10.3   Generalization Results (Metric 2)

We see in Table 7 that our Model 2 has better generalization capabilities using Metric 2 than other methods overall and for most specific generalization scenarios. In two of the generalization senarios, Model 2 was a close second behind our Model 1 and MV CCA. Recall that with Metric 1, our Model 2 was best in generalization across all scenarios.

Table 7: (Metric 2) Generalization Results (three KDTree databases, one for each fact type) K10 metric

| | SP | SPO | | | | | | Total |
|---|---|---|---|---|---|---|---|---|
| NumFacts for this case | 14448 | 10605 | 9313 | 4842 | 4673 | 1755 | 3133 | 48769 |
| Case | S≥15,P≥15 | SP≥15, O≥15 | PO≥15, S≥15 | SO≥15, P≥15 | SP≥15, PO≥15 | SO≥15, PO≥15 | SO≥15, SP≥15 | All |
| Model2 | **3.656** | 1.991 | 1.802 | **2.709** | 1.854 | **2.635** | **1.162** | **2.476** |
| Model1 | 3.041 | **2.066** | 1.546 | 2.346 | 1.891 | 2.248 | 1.009 | 2.205 |
| MV CCA | 2.199 | 1.382 | **1.907** | 1.320 | 1.480 | 1.372 | 0.737 | 1.686 |
| ESZSL | 1.098 | 0.258 | 0.214 | 0.211 | 0.165 | 0.204 | 0.201 | 0.479 |
| TACL15 (retrained) | 1.619 | 1.140 | 1.385 | 1.114 | 1.454 | 1.457 | 1.215 | 1.372 |
| TACL15 (COCO pretrained) | 0.121 | 0.067 | 0.040 | 0.051 | 0.050 | 0.000 | 0.038 | 0.070 |

## 11 Language View Retrieval Qualitative Results

Facts colored in red were never seen during training. Facts colored in blue have at least one training example.

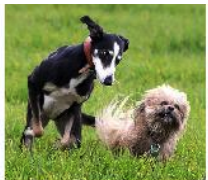

<S: dog, P:running, O:dog>, score = 0.75664
<S: dog, P:chasing, O:dog>, score = 0.72817
<S: dog, P:catching, O:ball>, score = 0.71931
<S: dog, P:holding, O:ball>, score = 0.70183
<S: dog, P:catching, O:stick>, score = 0.68052

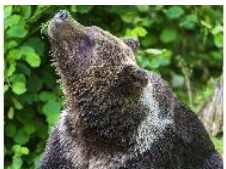

<S: bear, P:laying, O:head>, score = 0.68589
<S: bear, P:resting, O:head>, score = 0.67442
<S: bear, P:turns, O:face>, score = 0.61071
<S: bear, P:biting, O:neck>, score = 0.53843
<S: bear, P:taking, O:break>, score = 0.52448

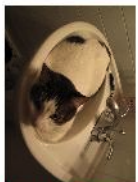

<S: cat, P:sitting, O:sink>, score = 0.71426
<S: cat, P:putting, O:inside>, score = 0.69014
<S: cat, P:lays, O:head>, score = 0.65617
<S: cat, P:resting, O:head>, score = 0.65588
<S: cat, P:touching, O:nose>, score = 0.63458

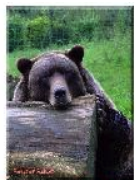

<S: bear, P:laying, O:head>, score = 0.67284
<S: bear, P:looking, O:upwards>, score = 0.53906
<S: bear, P:eating, O:fruit>, score = 0.49539
<S: elephant, P:walking, O:trees>, score = 0.33818
<S: giraffe, P:looks, O:tree>, score = 0.30462

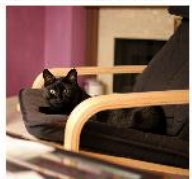

<S: cat, P:sitting, O:chair>, score = 0.68881
<S: cat, P:laying, O:chair>, score = 0.6532
<S: cat, P:putting, O:inside>, score = 0.64177
<S: cat, P:looking, O:door>, score = 0.61645
<S: cat, P:resting, O:head>, score = 0.58914

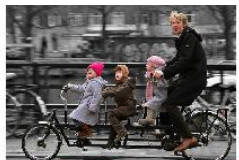

<S: men, P:riding, O:bike>, score = 0.84421
<S: people, P:riding, O:bike>, score = 0.83122
<S: person, P:riding, O:bike>, score = 0.78577
<S: people, P:ride, O:bike>, score = 0.76879
<S: men, P:riding, O:street>, score = 0.67821

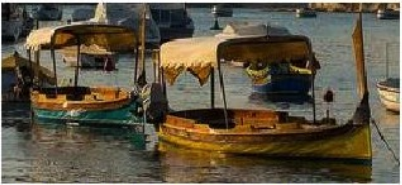

<S: boat, P:behind, O:boat>, score = 0.77582
<S: boat, P:beside, O:boat>, score = 0.76525
<S: boat, P:pulling, O:boat>, score = 0.75362
<S: boats, P:going, O:water>, score = 0.71488
<S: boats, P:over, O:water>, score = 0.70904

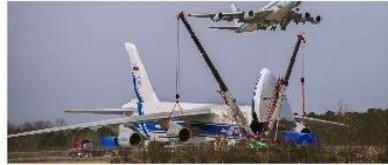

<S: airplane, P:flying, O:sky>, score = 0.71754
<S: planes, P:flying, O:sky>, score = 0.66272
<S: airplane, P:flying, O:beach>, score = 0.62434
<S: airplane, P:leaving, O:smoke>, score = 0.57553
<S: airplane, P:flying, O:style>, score = 0.57238

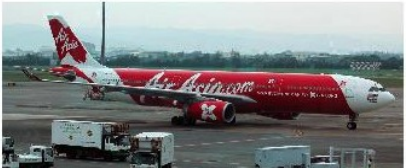

<S: airplane, P:above, O:truck>, score = 0.72659
<S: airplane, P:on, O:road>, score = 0.69693
<S: airplane, P:over, O:ground>, score = 0.67254
<S: airplane, P:on, O:ground>, score = 0.65618
<S: airplane, P:sits, O:front>, score = 0.65471

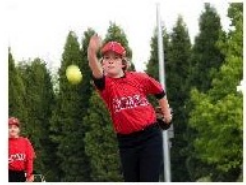

<S: boy, P:playing, O:ball>, score = 0.60118
<S: woman, P:throwing, O:ball>, score = 0.58973
<S: boy, P:wearing, O:ball>, score = 0.58424
<S: woman, P:hitting, O:ball>, score = 0.55619
<S: man, P:hitting, O:ball>, score = 0.55289

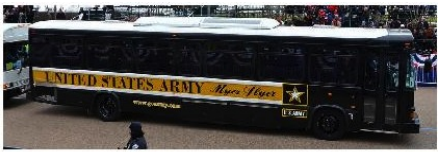

<S: bus, P:behind, O:bus>, score = 0.85044
<S: bus, P:parked, O:bus>, score = 0.83185
<S: bus, P:beside, O:bus>, score = 0.827
<S: bus, P:passing, O:bus>, score = 0.80861
<S: bus, P:near, O:bus>, score = 0.78307
<S: bus, P:says, O:bus>, score = 0.78229

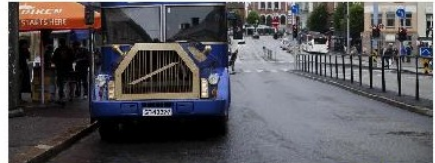

<S: bus, P:driving, O:street>, score = 0.7294
<S: bus, P:moving, O:street>, score = 0.71618
<S: bus, P:entering, O:street>, score = 0.70884
<S: bus, P:traveling, O:street>, score = 0.70281
<S: bus, P:on, O:road>, score = 0.67702

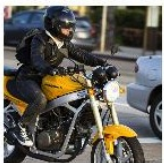

<S: man, P:riding, O:motorcycle>, score = 0.82711
<S: man, P:riding, O:bike>, score = 0.81556
<S: policeman, P:riding, O:motorcycle>, score = 0.73218
<S: man, P:behind, O:motorcycle>, score = 0.69656
<S: one, P:riding, O:bike>, score = 0.68641

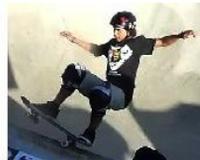

<S: man, P:riding, O:skateboard>, score = 0.80536
<S: man, P:jumping, O:skateboard>, score = 0.77956
<S: person, P:riding, O:skate>, score = 0.71684
<S: man, P:riding, O:ramp>, score = 0.58606
<S: man, P:riding, O:bull>, score = 0.58022
<S: man, P:riding, O:pair>, score = 0.56445

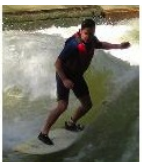

<S: man, P:riding, O:wave>, score = 0.77081
<S: man, P:riding, O:waves>, score = 0.75158
<S: man, P:jumping, O:skateboard>, score = 0.67344
<S: guy, P:riding, O:wave>, score = 0.65388
<S: man, P:riding, O:scooter>, score = 0.64177

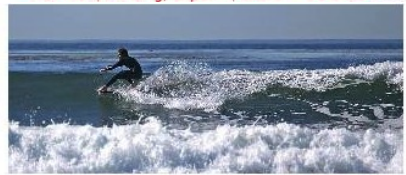

<S: man, P:riding, O:wave>, score = 0.90756
<S: man, P:riding, O:waves>, score = 0.87589
<S: guy, P:riding, O:wave>, score = 0.85144
<S: one, P:riding, O:wave>, score = 0.85065
<S: he, P:riding, O:wave>, score = 0.83461

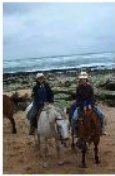

<S: people, P:riding, O:horse>, score = 0.90744
<S: people, P:riding, O:horses>, score = 0.88037
<S: person, P:riding, O:horse>, score = 0.87343
<S: men, P:riding, O:horse>, score = 0.82576
<S: he, P:riding, O:horse>, score = 0.82069

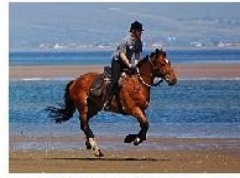

<S: person, P:riding, O:horse>, score = 0.9039
<S: man, P:riding, O:horse>, score = 0.90214
<S: someone, P:riding, O:horse>, score = 0.87873
<S: man, P:riding, O:horses>, score = 0.86064
<S: people, P:riding, O:horse>, score = 0.85644

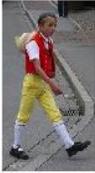

<S: boy, P:wearing, O:shirt>, score = 0.75567
<S: boy, P:wearing, O:jacket>, score = 0.71906
<S: boy, P:wearing, O:clothes>, score = 0.70289
<S: boy, P:wearing, O:clothing>, score = 0.68882
<S: boy, P:wearing, O:shoes>, score = 0.67941

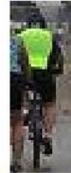

<S: person, P:riding, O:bike>, score = 0.71211
<S: one, P:walking, O:bike>, score = 0.62366
<S: people, P:riding, O:bike>, score = 0.58687
<S: athlete, P:riding, O:bike>, score = 0.58031
<S: man, P:walking, O:dog>, score = 0.57705

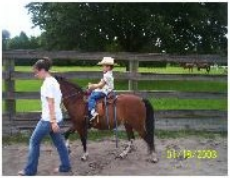

<S: people, P:riding, O:horse>, score = 0.80125
<S: men, P:riding, O:horse>, score = 0.79059
<S: woman, P:riding, O:horse>, score = 0.77016
<S: person, P:riding, O:horse>, score = 0.7628
<S: women, P:riding, O:horse>, score = 0.75578

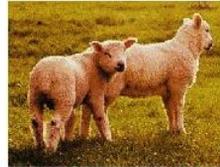

<S: sheep, P:standing, O:side>, score = 0.69933
<S: sheep, P:taking, O:rest>, score = 0.68388
<S: sheep, P:above, O:grass>, score = 0.68015
<S: sheep, P:beside, O:woman>, score = 0.62879
<S: cow, P:threw, O:fence>, score = 0.52324

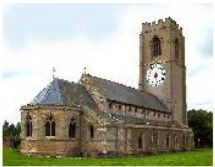

<S: tower, P:above, O:building>, score = 0.78785
<S: tower, P:covered, O:building>, score = 0.72547
<S: tower, P:on, O:building>, score = 0.71123
<S: roof, P:above, O:building>, score = 0.6842
<S: tower, P:has, O:roof>, score = 0.66889

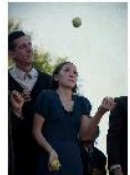

<S: woman, P:hitting, O:ball>, score = 0.63423
<S: woman, P:picking, O:ball>, score = 0.63169
<S: woman, P:bouncing, O:ball>, score = 0.61562
<S: woman, P:holding, O:kid>, score = 0.57083
<S: woman, P:juggling, O:balls>, score = 0.55341

## 12    Visual View Retrieval Qualitative Results

Facts colored in red were never seen during training. Facts colored in blue have at least one training example. Green boxes in the retrieved images indicate the images that were annotated by the query fact. It is easy to see that the method retrieves a lot of relevant examples for which it was not given credit, which opens the door to explore better metrics for Sherlock Problem. As illustrated in the experiments section, our large scale setting has hundreds of thousands of facts with the majority of them have one or few examples. The following examples show how our model managed to retrieve these examples to the top of the list given the fact in the language view.

### 12.1    Unseen Facts during training

<S: airplane, P:flying_over, O:ocean, A:>
<S: birds, P:eating_from, O:water, A:>

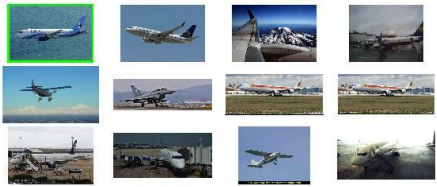
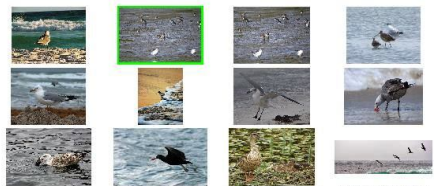

<S: boy, P:holding, O:toothbrushes, A:>
<S: broccoli, P:filling, O:bowl, A:>

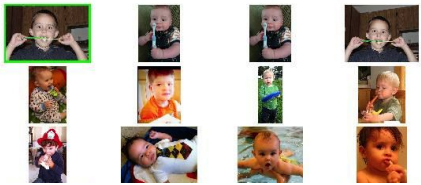
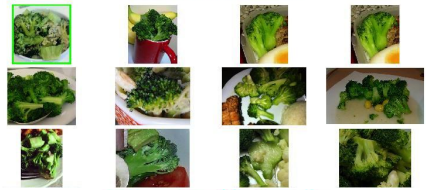

<S: dog, P:standing_on, O:truck, A:>
<S: drawer, P:next_to, O:oven, A:>

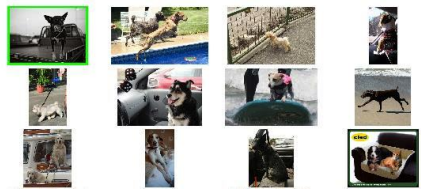
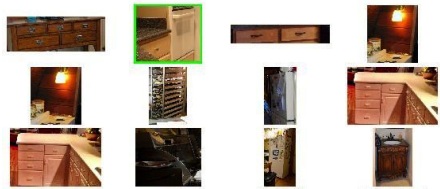

<S: elephant, P:nearing, O:food, A:>
<S: giraffe, P:walking, O:road, A:>

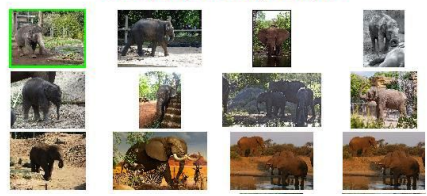
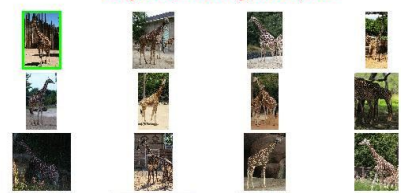

**<S: girl, P:stands_under, O:umbrella, A:>**

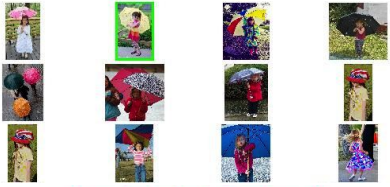

**<S: girl, P:using, O:racket, A:>**

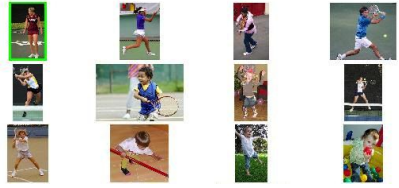

**<S: man, P:urinating_in, O:restroom, A:>**

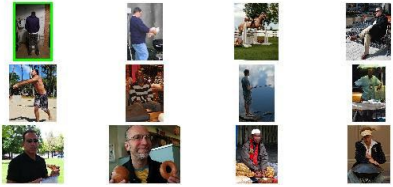

**<S: snow, P:on_top_of, O:mountains, A:>**

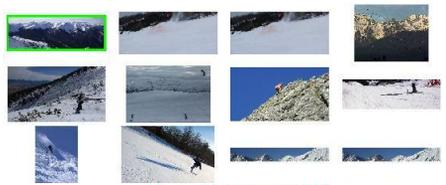

**<S: stove, P:left_of, O:sink, A:>**

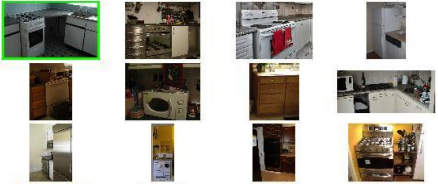

**<S: stove, P:left_of, O:sink, A:>**

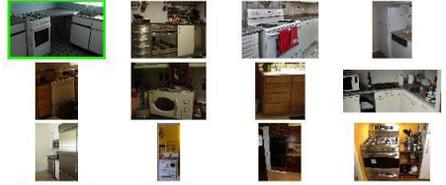

**<S: stove, P:left_of, O:sink, A:>**

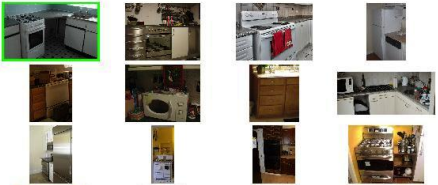

**<S: text, P:above, O:pictures, A:>**

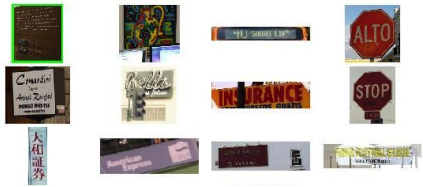

**<S: train, P:is_pulling_out_of, O:station, A:>**

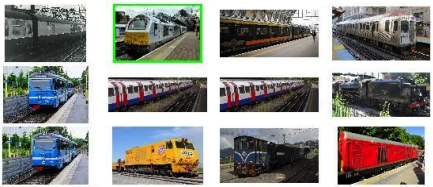

**<S: tree, P:behind, O:ramp, A:>**

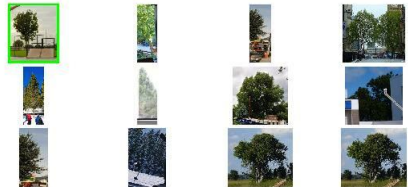

## 12.2   Seen Facts during training

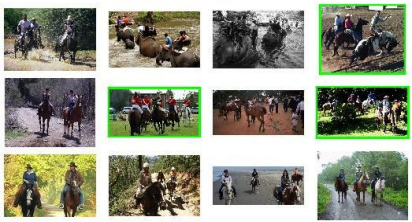

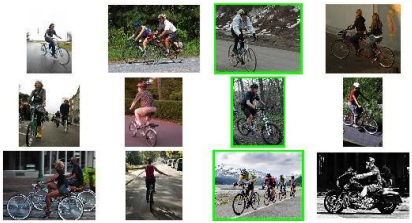

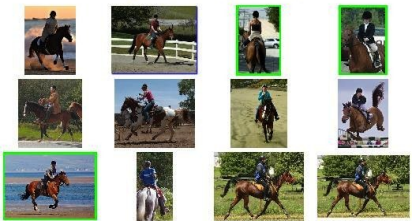

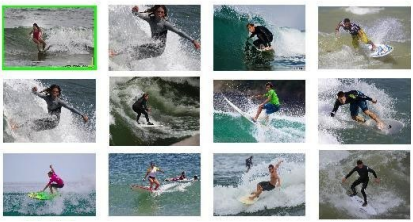

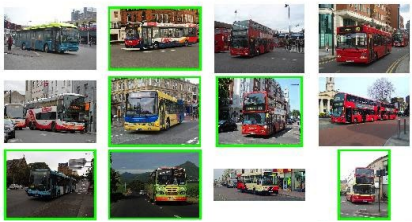

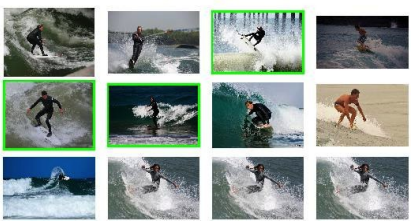

**<S: person, P:riding, O:bicycle, A:>**

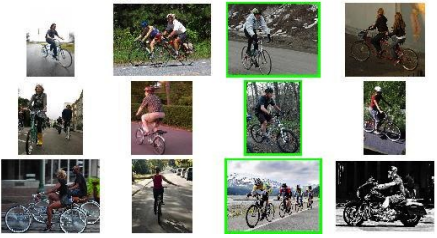

**<S: man, P:riding, O:skateboard, A:>**

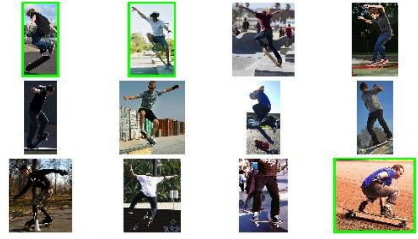

**<S: man, P:flying, O:kite, A:>**

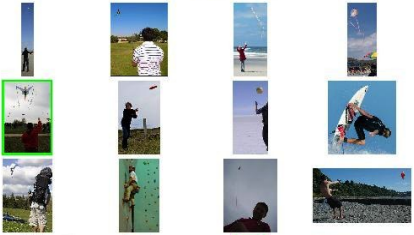

**<S: woman, P:flying, O:kite, A:>**

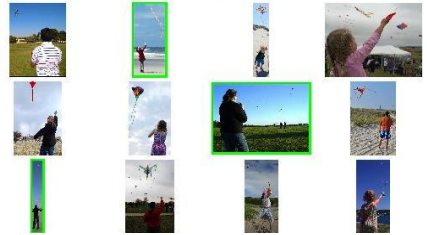

**<S: dog, P:lying_on, O:sofa, A:>**

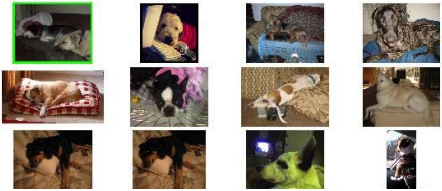

**<S: girl, P:carrying, O:luggage, A:>**

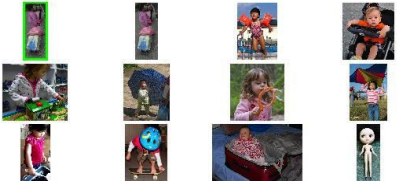

**<S: girl, P:holds, O:surfboard, A:>**

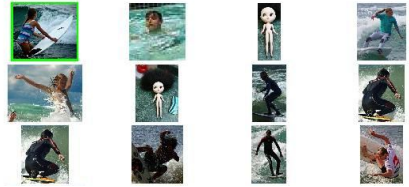

**<S: girl, P:walking, O:dog, A:>**

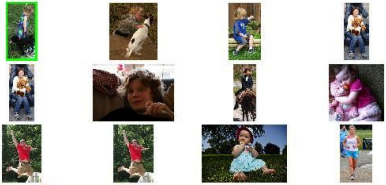

**<S: group, P:covered, O:mountain, A:>**

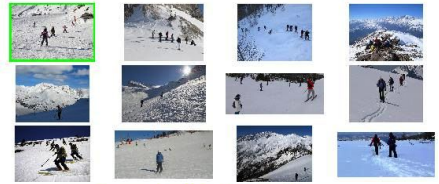

**<S: people, P:riding, O:surfboards, A:>**

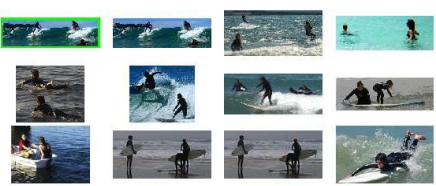

**<S: group, P:riding, O:wave, A:>**

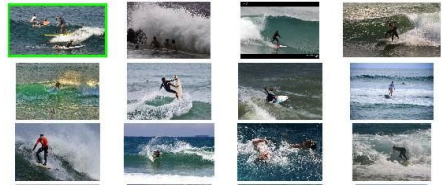

**<S: people, P:cutting, O:cake, A:>**

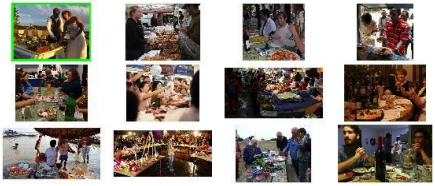

**<S: people, P:skiing, O:hill, A:>**

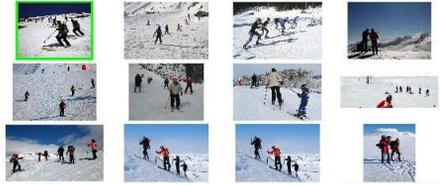

**<S: person, P:sitting_behind, O:person, A:>**

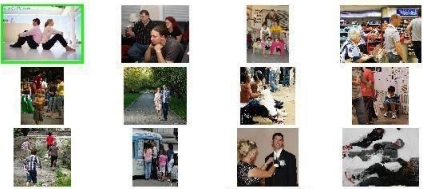

**<S: dogs, P:chase, O:dog, A:>**

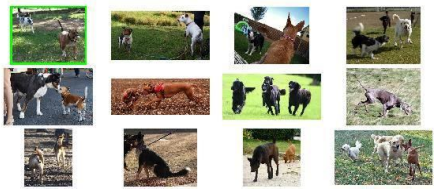

## 13  Qualitative Results for Language View Retrieval Generalization

In all the following figures, the number of training images are less than 5 examples, but the model shows some generalization cases that we discussed in Table 5 in the main paper and Table 2 in this document. In the following examples, the ground truth is on the top of the list.

Fig. 14: SPO≤5, PO≥15 and S≥15

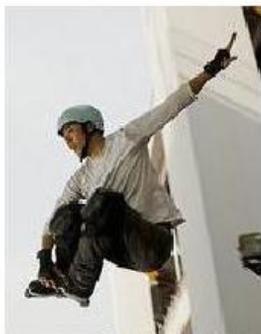

<S: man, P:jumping, O:skateboard>, score = 0.75385
<S: man, P:holding, O:shirt>, score = 0.57555
<S: man, P:wearing, O:pants>, score = 0.56939
<S: man, P:pulling, O:rope>, score = 0.55689
<S: man, P:wearing, O:helmet>, score = 0.5553

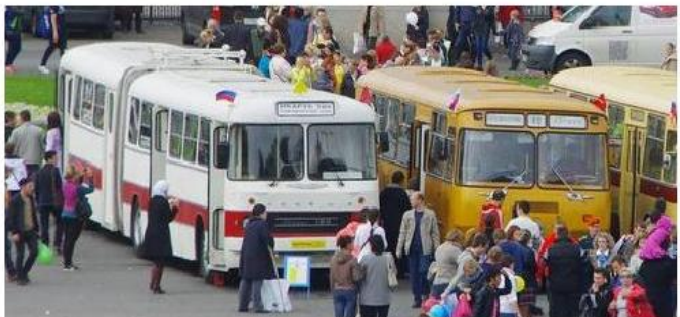

<S: bus, P:behind, O:bus>, score = 0.79708
<S: bus, P:parked, O:bus>, score = 0.7812
<S: bus, P:passing, O:bus>, score = 0.7633
<S: bus, P:over, O:road>, score = 0.74798
<S: bus, P:on, O:road>, score = 0.7465
<S: bus, P:near, O:bus>, score = 0.7438

Fig. 15: SPO≤5, SO≥15 and P≥15

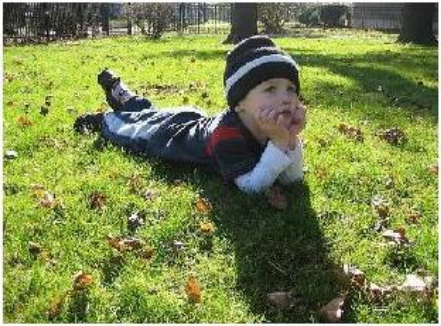

```
<S: boy, P:laying, O:grass>, score = 0.63062
<S: boy, P:on, O:grass>, score = 0.58784
<S: man, P:above, O:grass>, score = 0.49804
<S: boy, P:holding, O:fruit>, score = 0.49597
<S: man, P:jumping, O:snow>, score = 0.45806
<S: man, P:behind, O:fence>, score = 0.45472
```

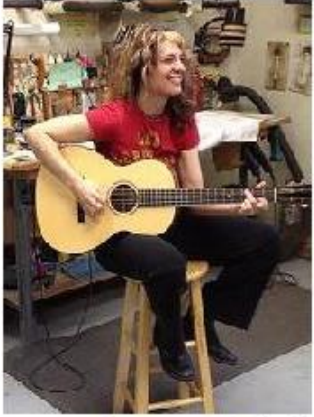

```
<S: woman, P:playing, O:violin>, score = 0.91379
<S: woman, P:playing, O:guitar>, score = 0.90077
<S: woman, P:playing, O:piano>, score = 0.89158
<S: woman, P:playing, O:cello>, score = 0.88553
<S: woman, P:plays, O:guitar>, score = 0.81954
<S: girl, P:playing, O:guitar>, score = 0.81102
<S: woman, P:plays, O:piano>, score = 0.80898
<S: lady, P:playing, O:guitar>, score = 0.80674
<S: girl, P:playing, O:flute>, score = 0.79454
<S: woman, P:holding, O:violin>, score = 0.77518
```

Fig. 16: SPO≤5, SP≥15 and O≥15

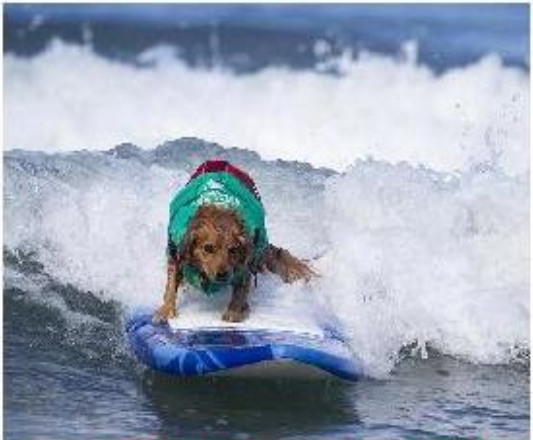

\<S: dog, P:riding, O:wave\>, score = 0.82598
\<S: someone, P:riding, O:waves\>, score = 0.73781
\<S: man, P:riding, O:waves\>, score = 0.71871
\<S: she, P:riding, O:wave\>, score = 0.70069
\<S: couple, P:riding, O:wave\>, score = 0.69202

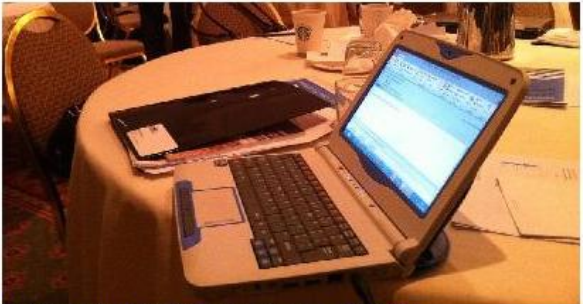

\<S: notebook, P:on, O:desk\>, score = 0.72809
\<S: laptop, P:has, O:screen\>, score = 0.72803
\<S: laptop, P:beside, O:monitor\>, score = 0.71566
\<S: notebook, P:on, O:wall\>, score = 0.62987
\<S: laptop, P:has, O:paper\>, score = 0.62961

Fig. 17: SPO≤5, SO≥15 and PO≥15

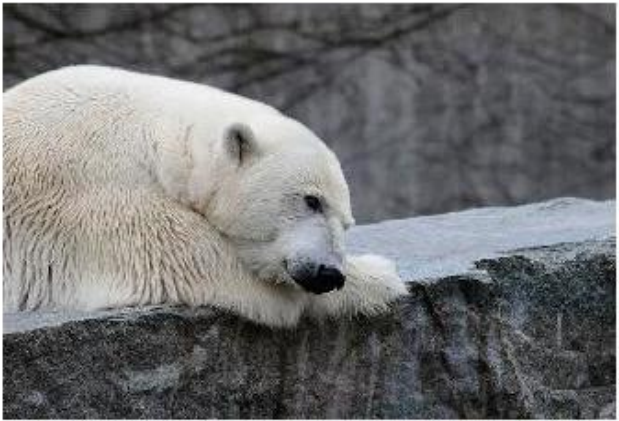

\<S: bear, P:laying, O:head>, score = 0.69381
\<S: bear, P:resting, O:head>, score = 0.66321
\<S: bear, P:sticking, O:head>, score = 0.65048
\<S: bear, P:turns, O:face>, score = 0.6375
\<S: bear, P:next_to, O:tree>, score = 0.61866
\<S: bear, P:sticking, O:tongue>, score = 0.58153

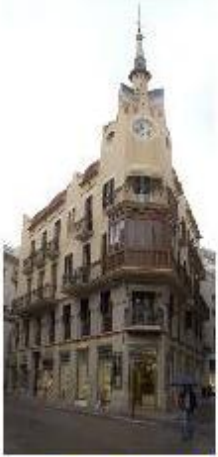

\<S: tower, P:above, O:building>, score = 0.78009
\<S: roof, P:above, O:building>, score = 0.67363
\<S: building, P:above, O:roof>, score = 0.66408
\<S: building, P:near, O:building>, score = 0.64902
\<S: building, P:has, O:tower>, score = 0.63818
\<S: roof, P:on, O:building>, score = 0.63178

Fig. 18: SPO≤5, SP≥15 and PO≥15

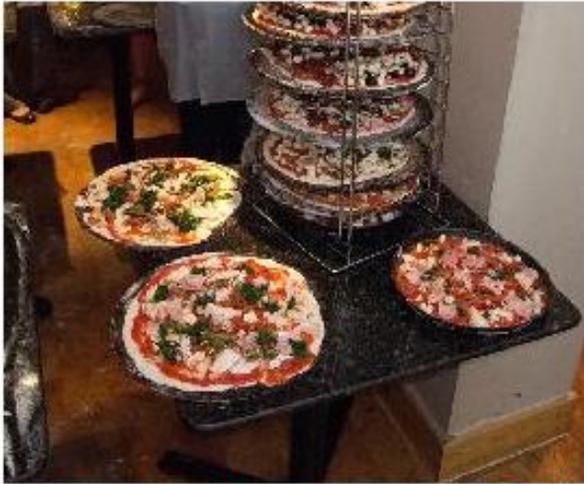

<S: table, P:under, O:bread>, score = 0.65614
<S: table, P:have, O:cake>, score = 0.64762
<S: table, P:under, O:cake>, score = 0.63287
<S: table, P:has, O:coffee>, score = 0.61875
<S: table, P:holding, O:trays>, score = 0.58758

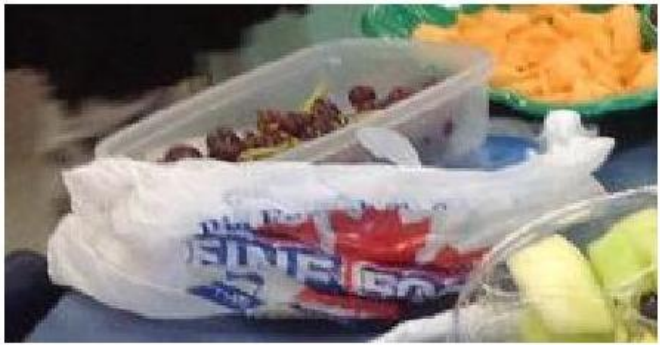

<S: bowl, P:has, O:fruit>, score = 0.54638
<S: cup, P:beside, O:cup>, score = 0.54186
<S: tomato, P:above, O:bread>, score = 0.53219
<S: egg, P:shaped, O:cake>, score = 0.53016
<S: cup, P:beside, O:plate>, score = 0.52929

Fig. 19: SPO≤5, SO≥15 and SP≥15

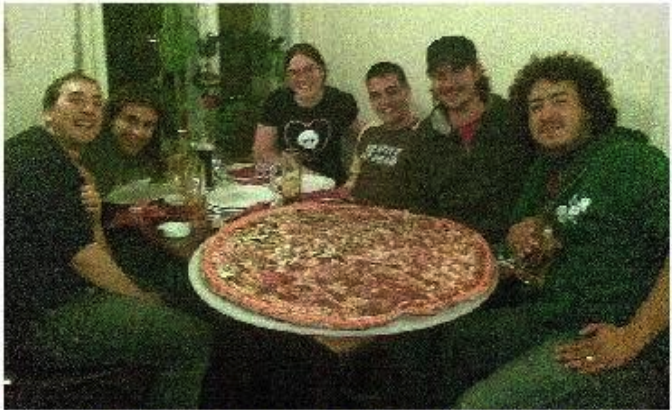

<S: people, P:eating, O:pizza>, score = 0.71083
<S: people, P:holding, O:pizza>, score = 0.70206
<S: people, P:eat, O:pizza>, score = 0.69191
<S: people, P:eating, O:food>, score = 0.66256
<S: people, P:enjoying, O:meal>, score = 0.66067
<S: people, P:having, O:meal>, score = 0.65211
<S: people, P:getting, O:food>, score = 0.64728
<S: people, P:sharing, O:pizza>, score = 0.64073